# Riffled Independence for Efficient Inference
# with Partial Rankings

**Jonathan Huang**                                                    JHUANG11@STANFORD.EDU
*James H. Clark Center*
*Stanford University, Stanford CA 94305, USA*

**Ashish Kapoor**                                                     AKAPOOR@MICROSOFT.COM
*Microsoft Research*
*One Microsoft Way*
*Redmond WA 98052-6399, USA*

**Carlos Guestrin**                                                   GUESTRIN@CS.CMU.EDU
*Gates Hillman Complex, Carnegie Mellon University,*
*5000 Forbes Avenue, Pittsburgh, PA 15213, USA*

## Abstract

Distributions over rankings are used to model data in a multitude of real world settings such as preference analysis and political elections. Modeling such distributions presents several computational challenges, however, due to the factorial size of the set of rankings over an item set. Some of these challenges are quite familiar to the artificial intelligence community, such as how to compactly represent a distribution over a combinatorially large space, and how to efficiently perform probabilistic inference with these representations. With respect to ranking, however, there is the additional challenge of what we refer to as *human task complexity* — users are rarely willing to provide a full ranking over a long list of candidates, instead often preferring to provide partial ranking information.

Simultaneously addressing all of these challenges — i.e., designing a compactly representable model which is amenable to efficient inference *and* can be learned using partial ranking data — is a difficult task, but is necessary if we would like to scale to problems with nontrivial size. In this paper, we show that the recently proposed *riffled independence* assumptions cleanly and efficiently address each of the above challenges. In particular, we establish a tight mathematical connection between the concepts of riffled independence and of partial rankings. This correspondence not only allows us to then develop efficient and exact algorithms for performing inference tasks using riffled independence based representations with partial rankings, but somewhat surprisingly, also shows that efficient inference is not possible for riffle independent models (in a certain sense) with observations which do not take the form of partial rankings. Finally, using our inference algorithm, we introduce the first method for learning riffled independence based models from partially ranked data.

## 1. Probabilistic Modeling of Ranking Data: Three Challenges

Rankings arise in a number of machine learning application settings such as preference analysis for movies and books (Lebanon & Mao, 2008) and political election analysis (Gormley & Murphy, 2007; Huang & Guestrin, 2010). In many of these problems, it is of great interest to build statistical models over ranking data in order to make predictions, form recommendations, discover latent trends and structure and to construct human-comprehensible data summaries.

Modeling distributions over rankings is a difficult problem, however, due to the fact that as the number of items being ranked increases, the number of possible rankings increases factorially. This combinatorial explosion forces us to confront three central challenges when dealing with rankings. First, we need to deal with *storage complexity* — how can we compactly represent a distribution over the space of rankings?[1] Then there is *algorithmic complexity* — how can we efficiently answer probabilistic inference queries given a distribution?

Finally, we must contend with what we refer to as *human task complexity*, which is a challenge stemming from the fact that it can be difficult to accurately elicit a full ranking over a large list of candidates from a human user; choosing from a list of $n!$ options is no easy task and users typically prefer to provide *partial information*. Take the American Psychological Association (APA) elections, for example, which allow their voters to rank order candidates from favorite to least favorite. In the 1980 election, there were five candidates, and therefore $5! = 120$ ways to rank those five candidates. Despite the small candidate list, most voters in the election preferred to only specify their top-$k$ favorite candidates rather than writing down full rankings on their ballots (see Figure 1). For example, roughly a third of voters simply wrote down their single favorite candidate in this 1980 election.

These three intertwined challenges of storage, algorithmic, and human task complexity are the central issues of probabilistic modeling for rankings, and models that do not efficiently handle all three sources of complexity have limited applicability. In this paper, we examine a flexible and intuitive class of models for rankings based on a generalization of probabilistic independence called *riffled independence*, proposed in our recent work (Huang & Guestrin, 2009, 2010). While our previous papers have focused primarily on representational (storage complexity) issues, we now concentrate on inference and incomplete observations (i.e., partial rankings), showing that in addition to storage complexity, riffle independence based models can efficiently address issues of algorithmic and human task complexity.

In fact the two issues of algorithmic and human task complexity are intricately linked for riffle independent models. By considering partial rankings, we give users more flexibility to provide as much or as little information as they care to give. In the context of partial ranking data, the most relevant inference queries also take the form of partial rankings. For example, we might want to predict a voter's second choice candidate given information about his first choice. One of our main contributions in this paper is to show that inference for such partial ranking queries can be performed particularly efficiently for riffle independent models.

The main contributions of our work are as follows:[2]

- We reveal a natural and fundamental connection between riffle independent models and partial rankings. In particular, we show that the collection of partial rankings over an item set form a complete characterization of the space of observations upon

---

| First Choice | Second Choice | Third Choice | Fourth Choice | Fifth Choice | # of votes |
|---|---|---|---|---|---|
| 5 | 3 | 4 | 2 | 1 | 37 |
| 3 | 4 | 5 | 1 | 2 | 30 |
| 1 | 2 | 3 | --- | --- | 27 |
| 3 | --- | --- | --- | --- | 1198 |
| 4 | 1 | 3 | --- | --- | 15 |
| 1 | 3 | --- | --- | --- | 302 |
| 3 | 1 | 2 | 5 | 4 | 186 |

Figure 1: Example partial ranking data (taken from the American Psychological Association election dataset, 1980)

which one can efficiently condition a riffle independent model. As a result, we show that when ranked items satisfy the riffled independence relationship, conditioning on partial rankings can be done efficiently, with running time $O(n \cdot |H|)$, where $|H|$ denotes the number of model parameters.

- We prove that, in a sense (which we formalize), it is impossible to efficiently condition riffle independent models on observations that do not take the form of partial rankings.

- We propose the first algorithm that is capable of efficiently estimating the structure and parameters of riffle independent models from heterogeneous collections of partially ranked data.

- We show results on real voting and preference data evidencing the effectiveness of our methods.

## 2. Riffled Independence For Rankings

A ranking, $\sigma$, of items in an item set $\Omega$ is a one-to-one mapping between $\Omega$ and a rank set $R = \{1, \ldots, n\}$ and is denoted using *vertical bar notation* as $\sigma^{-1}(1)|\sigma^{-1}(2)|\ldots|\sigma^{-1}(n)$. We say that $\sigma$ ranks item $i_1$ *before (or over)* item $i_2$ if the rank of $i_1$ is less than the rank of $i_2$. For example, $\Omega$ might be $\{Corn, Peas, Apples, Oranges\}$ and the ranking $Corn|Peas|Apples|Oranges$ encodes a preference of Corn over Peas which is in turn preferred over Apples and so on. The collection of all possible rankings of item set $\Omega$ is denoted by $S_\Omega$ (or just $S_n$ when $\Omega$ is implicit).

Since there are $n!$ rankings of $n$ items, it is intractable to estimate or even explicitly represent arbitrary distributions on $S_n$ without making structural assumptions about the underlying distribution. While there are many possible simplifying assumptions that one can make, we focus on an approach that we have proposed in recent papers (Huang & Guestrin, 2009, 2010) in which the ranks of items are assumed to satisfy an intuitive generalized notion of probabilistic independence known as *riffled independence*. In this paper, we argue that riffled independence assumptions are particularly effective in settings where one would like to make queries taking the form of partial rankings. In the remainder of this section, we review riffled independence.

The riffled independence assumption posits that rankings over the item set $\Omega$ are generated by independently generating rankings of smaller disjoint item subsets (say, $A$ and $B$) which partition $\Omega$, and piecing together a full ranking by interleaving (or riffle shuffling) these smaller rankings together. For example, to rank our item set of foods, one might first rank the vegetables and fruits separately, then interleave the two subset rankings to form a full ranking. To formally define riffled independence, we use the notions of *relative rankings* and *interleavings*.

**Definition 1** (Relative ranking map). Given a ranking $\sigma \in S_\Omega$ and any subset $A \subset \Omega$, the *relative ranking of items in $A$*, $\phi_A(\sigma)$, is a ranking, $\pi \in S_A$, such that $\pi(i) < \pi(j)$ if and only if $\sigma(i) < \sigma(j)$.

**Definition 2** (Interleaving map). Given a ranking $\sigma \in S_\Omega$ and a partition of $\Omega$ into disjoint sets $A$ and $B$, the *interleaving of $A$ and $B$ in $\sigma$* (denoted, $\tau_{AB}(\sigma)$) is a (binary) mapping from the rank set $R = \{1, \ldots, n\}$ to $\{A, B\}$ indicating whether a rank in $\sigma$ is occupied by $A$ or $B$. As with rankings, we denote the interleaving of a ranking by its vertical bar notation: $[\tau_{AB}(\sigma)](1)|[\tau_{AB}(\sigma)](2)|\ldots|[\tau_{AB}(\sigma)](n)$.

**Example 3.** *Consider a partitioning of an item set $\Omega$ into vegetables $A = \{Corn, \ Peas\}$ and fruits $B = \{Apples, Oranges\}$, as well as a full ranking over these four items: $\sigma = Corn|Oranges|Peas|Apples$. In this case, the relative ranking of vegetables in $\sigma$ is $\phi_A(\sigma) = Corn|Peas$ and the relative ranking of fruits in $\sigma$ is $\phi_B(\sigma) = Oranges|Apples$. The interleaving of vegetables and fruits in $\sigma$ is $\tau_{AB}(\sigma) = A|B|A|B$.*

**Definition 4** (Riffled Independence). Let $h$ be a distribution over $S_\Omega$ and consider a subset of items $A \subset \Omega$ and its complement $B$. The sets $A$ and $B$ are said to be *riffle independent* if $h$ decomposes (or factors) as:

$$h(\sigma) = m_{AB}(\tau_{AB}(\sigma)) \cdot f_A(\phi_A(\sigma)) \cdot g_B(\phi_B(\sigma)),$$

for distributions $m_{AB}$, $f_A$ and $g_B$, defined over interleavings and relative rankings of $A$ and $B$ respectively. In other words, $A$ and $B$ are riffle independent if the relative rankings of $A$ and $B$, as well as their interleaving are mutually independent. We refer to $m_{AB}$ as the *interleaving distribution* and $f_A$ and $g_B$ as the *relative ranking distributions*.

Riffled independence has been found to approximately hold in a number of real datasets (Huang & Guestrin, 2012). When such relationships can be identified in data, then instead of exhaustively representing all $n!$ ranking probabilities, one can represent just the factors $m_{AB}$, $f_A$ and $g_B$, which are distributions over smaller sets.

## 2.1 Hierarchical Riffle Independent Models

The relative ranking factors $f_A$ and $g_B$ are themselves distributions over rankings. To further reduce the parameter space, it is natural to consider hierarchical decompositions of item sets into nested collections of partitions (like hierarchical clustering). For example, Figure 2.1 shows a hierarchical decomposition where vegetables are riffle independent of fruits among the "healthy" foods, and these healthy foods are, in turn, riffle independent of the subset of desserts: $\{Doughnuts, M\&Ms\}$.

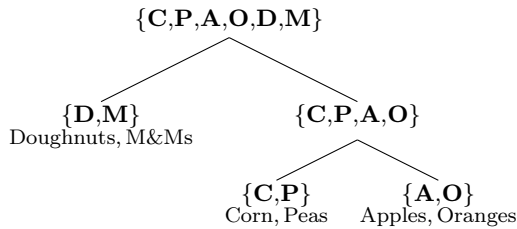

Figure 2: An example of a hierarchy over six food items.

For simplicity, we restrict consideration to binary hierarchies, defined as tuples of the form $H = (H_A, H_B)$, where $H_A$ and $H_B$ are either (1) null, in which case $H$ is called a *leaf*, or (2) hierarchies over item sets $A$ and $B$ respectively. In this second case, $A$ and $B$ are assumed to form a nontrivial partitioning of the item set.

**Definition 5.** We say that a distribution $h$ factors riffle independently with respect to a hierarchy $H = (H_A, H_B)$ if item sets $A$ and $B$ are riffle independent with respect to $h$, and both $f_A$ and $g_B$ factor riffle independently with respect to subhierarchies $H_A$ and $H_B$, respectively.

Like Bayesian networks, these hierarchies represent families of distributions obeying a certain set of (riffled) independence constraints and can be parameterized locally. To draw from such a model, one generates full rankings recursively starting by drawing rankings of the leaf sets, then working up the tree, sequentially interleaving rankings until reaching the root. The parameters of these hierarchical models are simply the interleaving and relative ranking distributions at the internal nodes and leaves of the hierarchy, respectively.

In general, the number of total parameters required to represent a hierarchical riffle independent model can (as with Bayesian networks) still scale exponentially in the number of items. For example, the number of interleavings of $p$ items with $n - p$ items is $\binom{n}{p}$. It is often the case however, that much fewer parameters are necessary. For example, *thin models* (Huang & Guestrin, 2012), in which the number of items factored out of the model at each stage of the hierarchy is never more than a small constant $k$, can always be represented with a (degree $k$) polynomial number of parameters. We will use $|H|$ to refer to the number of parameters necessary for representing a distribution which factors according to hierarchy $H$.

By decomposing distributions over rankings into small pieces (like Bayesian networks have done for other distributions), these hierarchical models allow for better interpretability, efficient probabilistic representation, low sample complexity, efficient MAP optimization, and, as we show in this paper, efficient inference.

**Example 6.** *In Figure 3(a), we reproduce the hierarchical structure that was learned using a fully ranked subset of the APA data consisting of 5000 training examples in Huang and Guestrin (2012). There were five candidates in the election: (1) William Bevan, (2) Ira Iscoe, (3) Charles Kiesler, (4) Max Siegle, and (5) Logan Wright (Marden, 1995). Strikingly, the structure that is learned using an algorithm (maximum likelihood) which knows nothing about the underlying politics of the APA, has leaf nodes which correspond exactly to the political coalitions that dominated the APA in the 1980 election — the research psychologists*

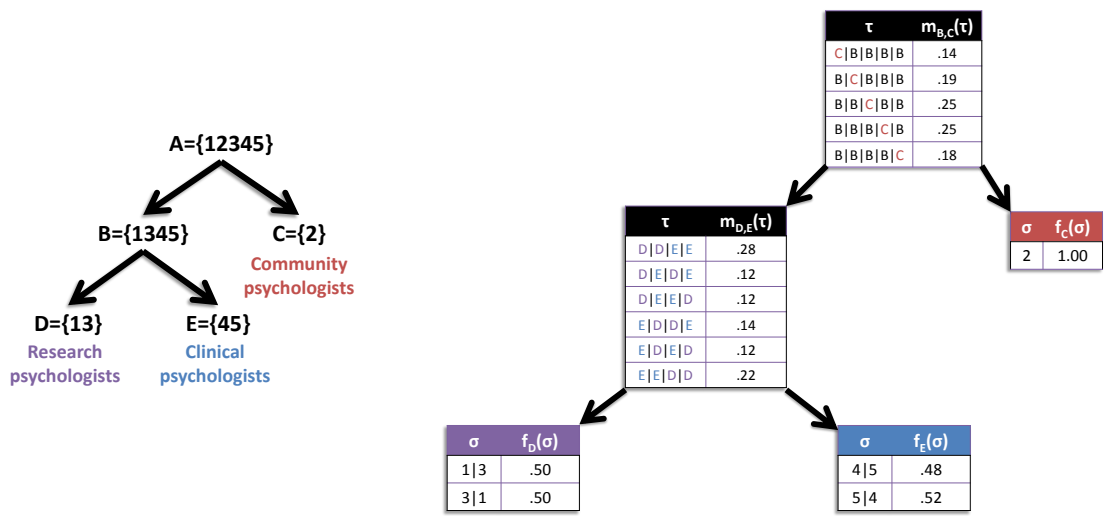

(a) Hierarchical structure learned via MLE using 5000 full rankings from the APA dataset.

(b) Riffle independent model parameters learned via MLE using 5000 full rankings from the APA dataset.

Figure 3: Example hierarchical model for the APA election. Candidates are enumerated as: (1) William Bevan, (2) Ira Iscoe, (3) Charles Kiesler, (4) Max Siegle, and (5) Logan Wright (Marden, 1995).

(candidates 1 and 3), the clinical psychologists (candidates 4 and 5), and the community psychologists (candidate 2).

In Figure 3(b), we plot the corresponding parameter distributions that are learned via maximum likelihood. There are three relative ranking distributions, each corresponding to a political party, as well as two interleaving distributions (one for the interleaving of research and clinical psychologists, and one for the interleaving of the community psychologist and all remaining candidates). Since each parameter distribution is constrained to sum to 1, there are a total of 11 free parameters.

## 2.2 Model Estimation

In this paper we estimate riffle independent models based on the methods introduced in our earlier work. Given the hierarchial structure of a model, the maximum likelihood parameter estimates of a hierarchical riffle independent model are straightforward to compute via frequency estimates. But how to estimate the correct structure of a model is a more challenging problem. The key insight lies in noticing that if two subsets $A$ and $B$ are riffle independent, then for any $i \in A$ and $j, k \in B$, the independence relation $\sigma(i) \perp (\sigma(j) < \sigma(k))$ must hold. Our structure learning algorithms operate by hunting for these 'tripletwise' independence relations within the data. We defer interested readers to the details in (Huang & Guestrin, 2012).

Note that in our earlier work, we assumed that our algorithms have access to a dataset consisting of i.i.d. full rankings provided by users. In the current work, we will relax our assumptions by allowing for users to provide partially ranked data. One assumption throughout, however, is that each user has a full ranking in mind over the items. In particular, our current work does not address the *incomplete ranking problem*, in which users might not have seen all of the items (we discuss possible extensions to the incomplete ranking setting in Section 9.

## 3. Decomposable Observations

Given a prior distribution, $h$, over rankings and an observation $\mathcal{O}$, Bayes rule tells us that the posterior distribution, $h(\sigma|\mathcal{O})$, is proportional to $L(\mathcal{O}|\sigma) \cdot h(\sigma)$, where $L(\mathcal{O}|\sigma)$ is the likelihood function. This operation of *conditioning* $h$ on an observation $\mathcal{O}$ is typically computationally intractable since it requires multiplying two $n!$ dimensional functions, unless one can exploit structural decompositions of the problem. In this section, we describe a decomposition for a certain class of likelihood functions over the space of rankings in which the observations are 'factored' into simpler parts. When an observation $\mathcal{O}$ is decomposable in this way, we show that one can efficiently condition a riffle independent prior distribution on $\mathcal{O}$. For simplicity in this paper, we focus primarily on *subset observations* whose likelihood functions encode membership with some subset of rankings in $S_n$.

**Definition 7** (Subset observations). A *subset observation* $\mathcal{O}$ is a binary observation whose likelihood is proportional to the indicator function of some subset of $S_n$ — i.e.,

$$L(\mathcal{O}|\sigma) = \left\{ \begin{array}{ll} 1 & \text{if } \sigma \in \mathcal{O} \\ 0 & \text{otherwise} \end{array} \right. .$$

As a running example, we will consider the class of *first place observations* throughout the chapter (we will consider far more general observation models in later sections). The first place observation $\mathcal{O} =$ "Corn is ranked first", for example, is associated with the collection of rankings placing the item Corn in first place ($\mathcal{O} = \{\sigma : \sigma(Corn) = 1\}$). We are interested in computing the posterior $h(\sigma|\mathcal{O} \in \mathcal{O})$. Thus in the first place scenario, we are given a voter's top choice and we would like to infer his preferences over the remaining candidates.

Given a partitioning of the item set $\Omega$ into two subsets $A$ and $B$, it is sometimes possible to *decompose* (or *factor*) a subset observation involving items in $\Omega$ into smaller subset observations involving $A$, $B$ and the interleavings of $A$ and $B$ independently. Such decompositions can often be exploited for efficient inference.

**Example 8.**

- *Consider the first place observation*

$$\mathcal{O} = \text{"Corn is ranked first"},$$

 *which can be decomposed into two independent observations — an observation on the relative ranking of Vegetables, and an observation on the interleaving of Vegetables and Fruits:*

- $\mathcal{O}_A$ = "Corn is ranked first among Vegetables",
- $\mathcal{O}_{A,B}$ = "First place is occupied by a Vegetable".

*To condition on $\mathcal{O}$ in this case, one updates the relative ranking distribution over Vegetables (A) by zeroing out rankings of vegetables which do not place Corn in first place, and updates the interleaving distribution by zeroing out interleavings which do not place a Vegetable in first place, then normalizes the resulting distributions.*

- *An example of a* nondecomposable *observation is the observation*

$$\mathcal{O} = \text{"Corn is in third place"}.$$

*To see that $\mathcal{O}$ does not decompose (with respect to Vegetables and Fruits), it is enough to notice that the interleaving of Vegetables and Fruits is not independent of the relative ranking of Vegetables. If, for example, an element $\sigma \in \mathcal{O}$ interleaves A (Vegetables) and B (Fruits) as $\tau_{AB}(\sigma) = A|B|A|B$, then since $\sigma(Corn) = 3$, the relative ranking of Vegetables is constrained to be $\phi_A(\sigma) = Peas|Corn$. Since the interleavings and relative rankings are not independent, we see that $\mathcal{O}$ cannot be decomposable.*

Formally, we use riffle independent factorizations to define decomposability with respect to a hierarchy $H$ of the item set.

**Definition 9** (Decomposability). Given a hierarchy $H$ over the item set, a subset observation $\mathcal{O}$ *decomposes* with respect to $H$ if its likelihood function $L(\mathcal{O}|\sigma)$ factors riffle independently with respect to $H$.

When subset observations and the prior decompose according to the same hierarchy, we can show (as in Example 8) that the posterior also decomposes.

**Proposition 10.** *Let $H$ be a hierarchy over the item set. Given a prior distribution $h$ and a subset observation $\mathcal{O}$ which both decompose with respect to $H$, the posterior distribution $h(\sigma|\mathcal{O})$ also factors riffle independently with respect to $H$.*

*Proof.* Denote the likelihood function corresponding to $\mathcal{O}$ by $L$ (in this proof, it does not matter that $\mathcal{O}$ is assumed to be a subset observation — the result holds for arbitrary likelihoods).

We use induction on the size of the item set $n = |\Omega|$. The base case $n = 1$ is trivially true. Next consider the general case where $n > 1$. The posterior distribution, by Bayes rule, can be written $h(\sigma|\mathcal{O}) \propto L(\sigma) \cdot h(\sigma)$. There are now two cases. If $H$ is a leaf node, then the posterior $h'$ trivially factors according to $H$, and we are done. Otherwise, $L$ and $h$ both factor, by assumption, according to $H = (H_A, H_B)$ in the following way:

$$L(\sigma) = m_L(\tau_{AB}(\sigma)) \cdot f_L(\phi_A(\sigma)) \cdot g_L(\phi_B(\sigma)), \text{ and } h(\sigma) = m_h(\tau_{AB}(\sigma)) \cdot f_h(\phi_A(\sigma)) \cdot g_h(\phi_B(\sigma)).$$

Multiplying and grouping terms, we see that the posterior factors as:

$$h(\sigma|\mathcal{O}) = [m_L \cdot m_h](\tau_{AB}(\sigma)) \cdot [f_L \cdot f_h](\phi_A(\sigma)) \cdot [g_L \cdot g_h](\phi_B(\sigma)).$$

To show that $h(\sigma|\mathcal{O})$ factors with respect to $H$, we need to demonstrate (by Definition 5) that the distributions $[f_L \cdot f_h]$ and $[g_L \cdot g_h]$ (after normalizing) factor with respect to $H_A$ and $H_B$, respectively. Since $f_L$ and $f_h$ both factor according to the hierarchy $H_A$ by assumption and $|A| < n$ since $H$ is not a leaf, we can invoke the inductive hypothesis to show that the posterior distribution, which is proportional to $f_L \cdot f_h$ must also factor according to $H_A$. Similarly, the distribution proportional to $g_L \cdot g_h$ must factor according to $H_B$. □

## 4. Complete Decomposability

The condition of Proposition 10, that the prior and observation must decompose with respect to *exactly* the same hierarchy, is a sufficient one for efficient inference, but it might at first glance seem so restrictive as to render the proposition useless in practice. To overcome this limitation of "hierarchy specific" decomposability, we explore a special family of observations (which we call *completely decomposable*) for which the property of decomposability does not depend specifically on a particular hierarchy, implying in particular that for these observations, efficient inference is *always* possible (provided that efficient representation of the prior distribution is also possible).

To illustrate how an observation can decompose with respect to multiple hierarchies over the item set, consider again the first place observation $\mathcal{O}$ ="Corn is ranked first". We argued in Example 8 that $\mathcal{O}$ is a decomposable observation. Notice however that decomposability for this particular observation *does not* depend on how the items are partitioned by the hierarchy. Specifically, if instead of Vegetables and Fruits, the sets $A = \{Corn, Apples\}$ and $B = \{Peas, Oranges\}$ are riffle independent, a similar decomposition of $\mathcal{O}$ would continue to hold, with $\mathcal{O}$ decomposing as an observation on the relative ranking of items in $A$ ("Corn is first among items in $A$"), and an observation on the interleaving of $A$ and $B$ ("First place is occupied by some element of $A$").

To formally capture this notion that an observation can decompose with respect to *arbitrary* underlying hierarchies, we define *complete decomposability*:

**Definition 11** (Complete decomposability). We say that a subset observation $\mathcal{O}$ is *completely decomposable* if it decomposes with respect to *every* possible hierarchy over the item set $\Omega$. We denote the collection of all possible completely decomposable (subset) observations as $\mathfrak{C}$. See Figure 4 for an illustration of the set $\mathfrak{C}$.

Conceptually, completely decomposable observations correspond to indicator functions that are "as riffle independent as possible". Complete decomposability is a *guarantee* for an observation $\mathcal{O}$ that one can always exploit any available factorized structure of the prior distribution in order to efficiently condition on $\mathcal{O}$.

**Proposition 12.** *Let $H$ be any binary hierarchy over the item set. Given a prior $h$ which factorizes with respect to $H$, and a completely decomposable observation $\mathcal{O}$, the posterior $h(\sigma|\mathcal{O})$ also decomposes with respect to $H$.*

*Proof.* Proposition 12 follows as a simple corollary to Proposition 10. □

**Example 13.** *The simplest example of a completely decomposable observation is the uniform observation $\mathcal{O}_{unif} = S_\Omega$, which includes all possible rankings and corresponds to a uniform indicator function $\delta_{unif}$ over rankings. Given any hierarchy $H$, $\delta_{unif}$ can be shown to decompose riffle independently with respect to $H$, where each factor is also uniform, and hence $\mathcal{O}_{unif}$ is completely decomposable.*

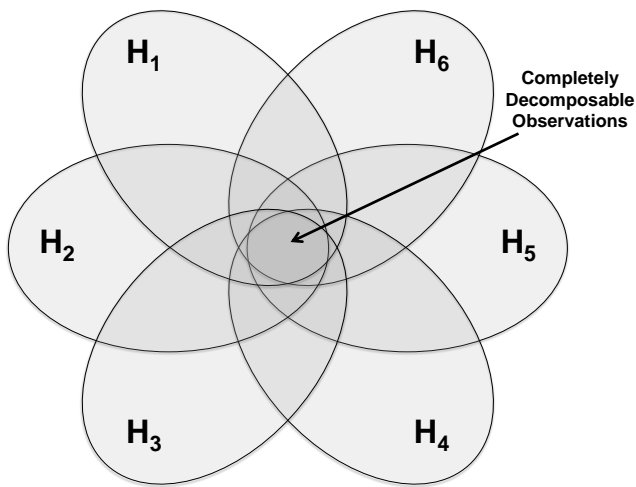

Figure 4: A diagram illustrating the collection of completely decomposable observations, $\mathfrak{C}$. Each shaded region (labeled $H_i$) above represents the family of subset observations over $S_n$ which decompose with respect to the hierarchy $H_i$. The collection $\mathfrak{C}$ can be seen as the intersection over all such shaded regions, and subset observations which lie inside of this intersection are ones for which conditioning can be performed in linear time (in the number of model parameters).

The uniform observation is of course not particularly interesting in the context of Bayesian inference, but on the other hand, given the stringent conditions in Definition 11, it is not obvious that nontrivial completely decomposable observations can even exist. Nonetheless, there do exist nontrivial examples (such as the first place observations), and in the next section, we exhibit a rich and general class of completely decomposable observations.

## 5. Complete Decomposability of Partial Ranking Observations

In this section we discuss the mathematical problem of fully characterizing the class of completely decomposable observations. Our main contribution in this section is to show that *completely decomposable observations correspond* precisely *to partial rankings of the item set*.

**Partial rankings.** We begin our discussion by introducing *partial rankings*, which allow for items to be tied with respect to a ranking $\sigma$ by 'dropping' verticals from the vertical bar representation of $\sigma$.

**Definition 14** (Partial ranking observation). Let $\Omega_1, \Omega_2, \ldots, \Omega_r$ be an ordered collection of subsets which partition $\Omega$ (i.e., $\cup_i \Omega_i = \Omega$ and $\Omega_i \cap \Omega_j = \emptyset$ if $i \neq j$). The *partial ranking observation*[3] corresponding to this partition is the collection of rankings which rank items

---

3. As remarked by Ailon (2007), we note that "The term *partial ranking* used here should not be confused with two other standard objects: (1) Partial order, namely, a reflexive, transitive anti-symmetric binary in $\Omega_i$ before items in $\Omega_j$ if $i < j$. We denote this partial ranking as $\Omega_1|\Omega_2|\ldots|\Omega_r$ and say that it has type $\gamma = (|\Omega_1|, |\Omega_2|, \ldots, |\Omega_r|)$. We denote the collection of all partial rankings (over $n$ items) as $\mathcal{P}$.

Each partial ranking as defined above can be viewed as a coset of the subgroup $S_\gamma = S_{\gamma_1} \times S_{\gamma_2} \times \cdots \times S_{\gamma_r}$. Given the type $\gamma$ and any full ranking $\pi \in S_\Omega$, there is only one partial ranking of type $\gamma$ containing $\pi$, thus we will therefore equivalently denote the partial ranking $\Omega_1|\Omega_2|\ldots|\Omega_r$ as $S_\gamma\pi$, where $\pi$ is any element of $\Omega_1|\Omega_2|\ldots|\Omega_r$. Note that this *coset notation* allows for multiple rankings $\sigma$ to refer to the same partial ranking $S_\gamma\sigma$.

The space of partial rankings as defined above captures a rich and natural class of observations. In particular, partial rankings encompass a number of commonly occurring special cases, which have traditionally been modeled in isolation, but in our work (as well as recent works such as Lebanon & Lafferty, 2003; Lebanon & Mao, 2008) can be used in a unified setting.

**Example 15.** *Partial ranking observations include:*

- (First place, or Top-1 observations): *First place observations correspond to partial rankings of type $\gamma = (1, n-1)$. The observation that "Corn is ranked first" can be written as Corn|Peas,Apples,Oranges.*

- (Top-$k$ observations): *Top-$k$ observations are partial rankings with type $\gamma = (1, \ldots, 1, n-k)$. These generalize the first place observations by specifying the items mapping to the first $k$ ranks, leaving all $n - k$ remaining items implicitly ranked behind. For example, the observation that "Corn is ranked first and Peas is ranked second" can be written as Corn|Peas|Apples,Oranges.*

- (Desired/less desired dichotomy): *Partial rankings of type $\gamma = (k, n-k)$ correspond to a subset of $k$ items being preferred or desired over the remaining subset of $n - k$ items. For example, partial rankings of type $(k, n-k)$ might arise in* approval voting *in which voters mark the subset of "approved" candidates, implicitly indicating disapproval of the remaining $n - k$ candidates.*

- (Ratings): *Finally, partial rankings can come in the form of rating data where, for example, restaurants are rated as, $\star$, $\star\star$, or $\star\star\star$. A corresponding partial ranking would thus "tie" restaurants that are rated with the same number of stars, while ranking restaurants with more stars above restaurants with fewer stars.*

- (Trivial observations): *Partial rankings of type $\gamma = (n)$ refer to trivial observations whose likelihood functions are uniform on the entire space of rankings, $S_\Omega$. The trivial observation for rankings of the item set $\Omega = \{Corn, Peas, Apples\}$, for example, can simply be written simply as Corn, Peas, Apples.*

To show how partial ranking observations decompose, we will exhibit an explicit factorization with respect to a hierarchy $H$ over items. For simplicity, we begin by considering the single layer case, in which the items are partitioned into two leaf sets $A$ and $B$. Our factorization depends on the following notions of *consistency* of relative rankings and interleavings with a partial ranking.

---

relation; and (2) A ranking of a subset of $\Omega$ [which we discuss in Section 9 as *incomplete rankings*]. In search engines, for example, although only the top-$k$ elements of $\Omega$ are returned, the remaining $n - k$ are implicitly assumed to be ranked behind [and therefore, search engines return partial rankings]."

**Definition 16** (Restriction consistency). Given a partial ranking $S_\gamma \pi = \Omega_1 | \Omega_2 | \ldots | \Omega_r$ and any subset $A \subset \Omega$, we define the *restriction* of $S_\gamma \pi$ to $A$ as the partial ranking on items in $A$ obtained by intersecting each $\Omega_i$ with $A$. Hence the restriction of $S_\gamma \pi$ to $A$ is:

$$[S_\gamma \pi]_A = \Omega_1 \cap A | \Omega_2 \cap A | \ldots | \Omega_r \cap A.$$

Given a ranking, $\sigma_A$ of items in $A$, we say that $\sigma_A$ is *consistent* with the partial ranking $S_\gamma \pi$ if $\sigma_A$ is a member of the restriction of $S_\gamma \pi$ to $A$, $[S_\gamma \pi]_A$.

**Definition 17** (Interleaving consistency). Given an interleaving $\tau_{AB}$ of two sets $A, B$ which partition $\Omega$, we say that $\tau_{AB}$ is *consistent* with a partial ranking $S_\gamma \pi = \Omega_1 | \ldots | \Omega_r$ (with type $\gamma$) if the first $\gamma_1$ entries of $\tau_{AB}$ contain the same number of As and Bs as $\Omega_1$, and the second $\gamma_2$ entries of $\tau_{AB}$ contain the same number of As and Bs as $\Omega_2$, and so on. Given a partial ranking $S_\gamma \pi$, we denote the collection of consistent interleavings as $[S_\gamma \pi]_{AB}$.

For example, consider the partial ranking

$$S_\gamma \pi = Corn, Apples | Peas, Oranges,$$

which places a single vegetable and a single fruit in the first two ranks, and a single vegetable and a single fruit in the last two ranks. Alternatively, $S_\gamma \pi$ partially specifies an interleaving $AB | AB$. The full interleavings $A | B | B | A$ and $B | A | B | A$ are consistent with $S_\gamma \pi$ (by dropping vertical lines) while $A | A | B | B$ is *not* consistent (since it places two vegetables in the first two ranks).

Using the notions of consistency with a partial ranking, we show that partial ranking observations are decomposable with respect to any binary partitioning (i.e., single layer hierarchy) of the item set.

**Proposition 18** (Single layer hierarchy). *For any partial ranking observation $S_\gamma \pi$ and any binary partitioning of the item set $(A, B)$, the indicator function of $S_\gamma \pi$, $\delta_{S_\gamma \pi}$, factors riffle independently as:*

$$\delta_{S_\gamma \pi}(\sigma) = m_{AB}(\tau_{AB}(\sigma)) \cdot f_A(\phi_A(\sigma)) \cdot g_B(\phi_B(\sigma)), \tag{5.1}$$

*where the factors $m_{AB}$, $f_A$ and $g_B$ are the indicator functions for consistent interleavings and relative rankings, $[S_\gamma \pi]_{AB}$, $[S_\gamma \pi]_A$ and $[S_\gamma \pi]_B$, respectively.*

The single layer decomposition of Proposition 18 can be turned into a recursive decomposition for partial ranking observations over arbitrary binary hierarchies, which establishes our main result. In particular, given a partial ranking $S_\gamma \pi$ and a prior distribution which factorizes according to a hierarchy $H$, we first condition the topmost interleaving distribution by zeroing out all parameters corresponding to interleavings which are not consistent with $S_\gamma \pi$, and normalizing the distribution. We then need to condition the subhierarchies $H_A$ and $H_B$ on relative rankings of $A$ and $B$ which are consistent with $S_\gamma \pi$, respectively. Since these consistent sets, $[S_\gamma \pi]_A$ and $[S_\gamma \pi]_B$, are partial rankings themselves, the same algorithm for conditioning on a partial ranking can be applied recursively to each of the subhierarchies $H_A$ and $H_B$. To be precise, we show that:

**Theorem 19.** *Every partial ranking is completely decomposable ($\mathcal{P} \subset \mathfrak{C}$).*

PRCONDITION (Prior $h_{prior}$, Hierarchy $H$, Observation $S_\gamma \pi = \Omega_1 | \Omega_2 | \ldots | \Omega_r$)

    **if** *isLeaf(H)* **then**
        **forall** $\sigma$ **do**

$$h_{post}(\sigma) \leftarrow \left\{ \begin{array}{ll} h_{prior}(\sigma) & \text{if } \sigma \in S_\gamma \pi \\ 0 & \text{otherwise} \end{array} \right. ;$$

        NORMALIZE $(h_{post})$ ;
        **return** $(h_{post})$;
    **else**
        **forall** $\tau$ **do**

$$m_{post}(\tau) \leftarrow \left\{ \begin{array}{ll} m_{prior}(\tau) & \text{if } \tau \in [S_\gamma \pi]_{AB} \\ 0 & \text{otherwise} \end{array} \right. ;$$

        NORMALIZE $(m_{post})$ ;
        $f(\phi_A) \leftarrow$PRCONDITION $(f_{prior}, H_A, [S_\gamma \pi]_A)$ ;
        $g(\phi_B) \leftarrow$PRCONDITION $(g_{prior}, H_B, [S_\gamma \pi]_B)$ ;
        **return** $(m_{post}, f_{post}, g_{post})$;

**Algorithm 1**: Pseudocode for PRCONDITION, an algorithm for recursively conditioning a hierarchical riffle independent prior distribution on partial ranking observations. See Definitions 16 and 17 for $[S_\gamma \sigma]_A$, $[S_\gamma \sigma]_B$, and $[S_\gamma \sigma]_{AB}$. The runtime of PRCONDITION is $O(n \cdot |H|)$, where $|H|$ is the number of model parameters. Input: All parameter distributions of the prior $h_{prior}$ represented in explicit tabular form, and an observation $S_\gamma \pi$ in the form of a partial ranking. Output: All parameter distributions of the posterior $h_{post}$ represented in explicit tabular form.

Since the proof of Theorem 19 is fairly straight forward given the form of the factorization (Equation 5.1), it is deferred to the Appendix. As a consequence of Theorem 19 and Proposition 12, conditioning on partial ranking observations can be performed efficiently. See Algorithm 1 for details on our recursive conditioning algorithm.

What is the running time complexity of conditioning on a partial ranking? The recursion of Algorithm 1 operates on each parameter distribution once, setting the probabilities of the interleavings or relative rankings in each such distribution to either zero or not, then normalizing. To decide whether to zero out a probability or not, one must check a partial ranking for consistency against either an interleaving or relative ranking, which requires at most $O(n)$ time. Therefore, in total, Algorithm 1 requires $O(n \cdot |H|)$ time, where $|H|$ is the total number of model parameters. Notice that the complexity of conditioning depends linearly on the complexity of the prior — whenever the prior distribution can be compactly represented, efficient inference for partial ranking observations is also possible. As we have stated in Section 2, $|H|$ can in general scale exponentially in $n$, but for *thin chain models*, in which the number of items factored out of the model at each stage is never more than a small constant $k$, verifying interleaving or relative ranking consistency can be performed in constant time, implying that the conditioning operation is linear in the number of model parameters, and guaranteed to be polynomial in $n$.

**Example 20.** *In this example, we consider conditioning the APA distribution from Example 6 on the observation $\mathcal{O}$ that "Candidate 3 is ranked in first place," which can also be represented as the partial ranking $\mathcal{O} = 3|1, 2, 4, 5$. Recall that candidate 3 was Charles Kiesler, who was a research psychologist.*

*In Figure 5(a) we show again the structure and parameters of the prior distribution for the APA election data, highlighting in particular the interleavings and relative rankings which*

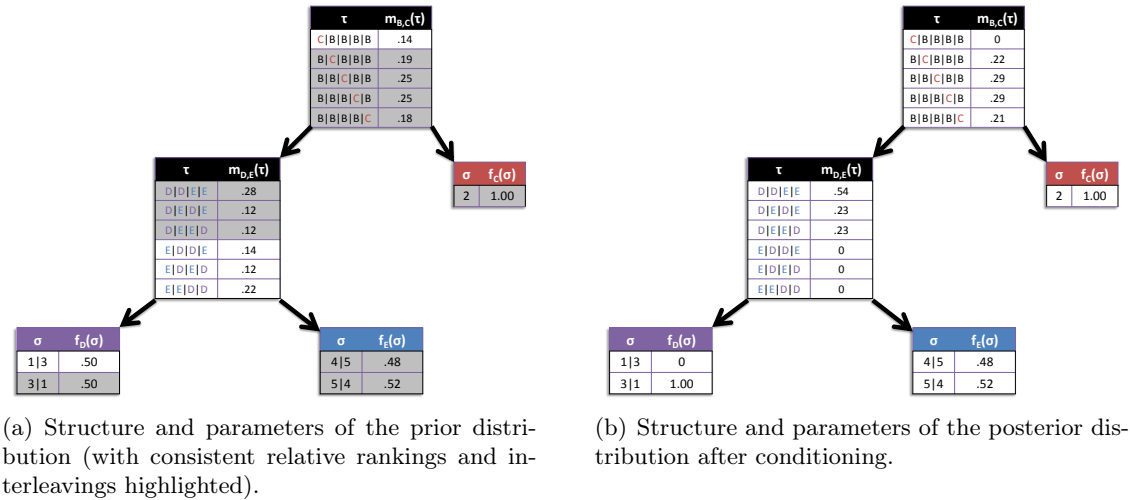

(a) Structure and parameters of the prior distribution (with consistent relative rankings and interleavings highlighted).

(b) Structure and parameters of the posterior distribution after conditioning.

Figure 5: Example of conditioning the APA hierarchy (from Example 6) on the first place observation that "Candidate 3 is ranked in first place".

are consistent with $\mathcal{O}$. For example, of the possible interleavings of research psychologists ($D$) with clinical psychologists ($E$), the interleavings that are consistent with $\mathcal{O}$ are those which rank a research psychologist first among the research and clinical psychologists. There are therefore only three consistent interleavings: $D|D|E|E$, $D|E|D|E$, and $D|E|E|D$.

Conditioning on $\mathcal{O}$ sets all relative rankings and interleavings which are *not* consistent with $\mathcal{O}$ to zero and normalizes each resulting parameter distribution. The resulting riffle independent representation of the posterior distribution is shown in Figure 5(b).

## 5.1 An Impossibility Result

It is interesting to consider what completely decomposable observations exist beyond partial rankings. One of our main contributions is to show that there are no such observations.

**Theorem 21** (Converse of Theorem 19). *Every completely decomposable observation takes the form of a partial ranking ($\mathfrak{C} \subset \mathcal{P}$).*

Together, Theorems 19 and 21 form a significant insight into the nature of rankings, showing that the notions of partial rankings and riffled independence are deeply connected. In fact, our result shows that it is even possible to define partial rankings via complete decomposability!

As a practical matter, Theorem 21 shows that there is no algorithm based on simple multiplicative updates to the parameters which can exactly condition on observations which do *not* take the form of partial rankings. The computational complexity of conditioning on observations which are not partial rankings remains open. We conjecture that approximate inference approaches may be necessary for efficiently handling more complex observations.

### 5.2 Proof of the Impossiblity Result (Theorem 21)

We now turn to proving Theorem 21. Since this proof is significantly longer and less obvious than the proof for its converse (Theorem 19), we sketch the main ideas that drive the proof here and refer interested readers to details in the Appendix.

Recall that the definition of the *linear span* of a set of vectors in a vector space is the intersection of all linear subspaces containing that set of vectors. To prove Theorem 21, we introduce analogous concepts of the *span* of a set of rankings.

**Definition 22** (RSPAN and PSPAN). Let $X \subset S_n$ be any collection of rankings. We define PSPAN($X$) to be the intersection of all partial rankings containing $X$. Similarly, we define RSPAN($X$) to be the intersection of all completely decomposable observations containing $X$. More formally,

$$\text{PSPAN}(X) = \bigcap_{S_\gamma \sigma : X \subset S_\gamma \sigma} S_\gamma \sigma, \quad \text{and} \quad \text{RSPAN}(X) = \bigcap_{\mathcal{O} : X \subset \mathcal{O}, \, \mathcal{O} \in \mathfrak{C}} \mathcal{O}.$$

For example, if $X = \{Corn|Peas|Apples, Apples|Peas|Corn\}$, it can be checked that the *only* partial ranking of all three items containing both items of $X$ is the entire set itself. Thus PSPAN($X$) = $Corn, Peas, Apples$.

Our proof strategy is to establish two claims: (1) that the PSPAN of any set is always a partial ranking, and (2) that in fact, the RSPAN and PSPAN of a set $X$ are exactly the same sets. Since claim (1) is a fact about partial rankings and does not involve riffled independence, we defer all related proofs to the Appendix. Thus we have:

**Lemma 23.** *For any $X \subset S_n$, PSPAN($X$) is a partial ranking.*

*Proof.* See Appendix. □

The following discussion will instead sketch a proof of claim (2). We first show, however, that Theorem 21 must hold if it is indeed true that claims (1) and (2) hold.

*Proof. (of Theorem 21):* Given some $\mathcal{O} \in \mathfrak{C}$, we want to show that $\mathcal{O} \in \mathcal{P}$. By claim (2), RSPAN($\mathcal{O}$) = PSPAN($\mathcal{O}$). Since $\mathcal{O}$ is an element of $\mathfrak{C}$, however, we also have that $\mathcal{O} = $ RSPAN($\mathcal{O}$), and thus that $\mathcal{O} = $ PSPAN($\mathcal{O}$). Finally Lemma 23 (claim (2)) guarantees that PSPAN($\mathcal{O}$) is a partial ranking, and so we conclude that $\mathcal{O} \in \mathcal{P}$. □

We now proceed to establish the claim that RSPAN($X$) = PSPAN($X$). The following proposition lists several basic properties of the RSPAN that we will use in several of the proofs. They all follow directly from definition so we do not write out the proofs.

**Proposition 24.**

I. *(Monotonicity) For any $X$, $X \subset$ RSPAN($X$).*

II. *(Subset preservation) For any $X, X'$ such that $X \subset X'$, RSPAN($X$) $\subset$ RSPAN($X'$).*

III. *(Idempotence) For any $X$, RSPAN(RSPAN($X$)) = RSPAN($X$).*

One inclusion of our proof that RSPAN($X$) = PSPAN($X$) follows directly from the fact that $\mathcal{P} \subset \mathfrak{C}$ (Theorem 19):

---

FORMPSPAN($X$)
    $X_0 \leftarrow X$; $t \leftarrow 0$;
    **while** $\exists S_\gamma \pi, S_{\gamma'} \pi' \in X_t$ *which disagree on the relative ordering of items* $a_1, a_2$ **do**
        $X_t \leftarrow \emptyset$ ;
        **foreach** $S_\gamma \sigma \in X_t$ **do**
            Add any partial ranking obtained by deleting a vertical bar from $S_\gamma \sigma$ between items
            $a_1$ and $a_2$ to $X_t$;
        $t \leftarrow t + 1$;
    **return** (any element of $X_t$) ;

---

**Algorithm 2**: Pseudocode for computing PSPAN($X$). FORMPSPAN($X$) takes a set of partial rankings (or full rankings) $X$ as input and outputs a partial ranking. This algorithm iteratively deletes vertical bars from elements of $X$ until they are in agreement. Note that it is not necessary to keep track of $t$, but we do so here to ease notation in the proofs. Nor is this algorithm the most direct way of computing PSPAN($X$), but again, it simplifies the proof of our main theorem.

**Lemma 25.** *For any subset of orderings, $X$,* RSPAN($X$) $\subset$ PSPAN($X$).

*Proof.* Fix a subset $X \subset S_n$ and let $\pi$ be any element of RSPAN($X$). We would like to show $\pi$ to be an element of PSPAN($X$). Consider any partial ranking $S_\gamma \sigma \in \mathcal{P}$ which covers $X$ (i.e., $\sigma' \in S_\gamma \sigma$ for all $\sigma' \in X$). We want to see that $\pi \in S_\gamma \sigma$. By Theorem 19, $\mathcal{P} \in \mathfrak{C}$, and therefore, $S_\gamma \sigma \in \mathfrak{C}$. Since $\pi \in$ RSPAN($X$), and $\sigma' \in S_\gamma \sigma$ for all $\sigma' \in X$, we conclude, by definition of RSPAN, that $\pi \in S_\gamma \sigma$. Since this holds for any partial ranking covering $X$, $\pi \in$ PSPAN($X$). $\qquad\square$

What remains is the task of establishing the reverse inclusion:

**Proposition 26.** *For any subset of orderings, $X$,* RSPAN($X$) $\supset$ PSPAN($X$).

To prove Proposition 26, we consider the problem of computing the partial ranking span (PSPAN) of a given set of rankings $X$. In Algorithm 2, we show a simple procedure based on iteratively finding rankings in $X$ which disagree on the pairwise ranking of two items, and replacing those rankings by a partial ranking in which a vertical bar between those two elements have been removed. We show that this algorithm provably outputs the correct result.

**Proposition 27.** *Given a set of rankings $X$ as input, Algorithm 2 outputs* PSPAN($X$).

*Proof.* See Appendix. $\qquad\square$

As a final step before being able to prove Proposition 26, we prove the following two technical lemmas which relate the computation of the PSPAN in Algorithm 2 to riffled independence, and really form the heart of our argument. In particular, for a completely decomposable observation $\mathcal{O} \in \mathfrak{C}$, Lemma 28 below shows how a ranking contained in $\mathcal{O}$ can "force" other rankings to also be contained in $\mathcal{O}$.

**Lemma 28.** *Let $\mathcal{O} \in \mathfrak{C}$ and suppose there exist $\pi_1, \pi_2 \in \mathcal{O}$ which disagree on the relative ranking of items $i, j \in \Omega$. Then the ranking obtained by swapping the relative ranking of items $i, j$ within any $\pi_3 \in \mathcal{O}$ must also be contained in $\mathcal{O}$.*

*Proof.* Let $h$ be the indicator distribution corresponding to the observation $\mathcal{O}$. We will show that swapping the relative ranking of items $i, j$ in $\pi_3$ will result in a ranking which is assigned nonzero probability by $h$, thus showing that this new ranking is contained in $\mathcal{O}$.

Let $A = \{i, j\}$ and $B = \Omega \backslash A$. Since $\mathcal{O} \in \mathfrak{C}$, $h$ must factor riffle independently according to the partition $(A, B)$. Thus,

$$h(\pi_1) = m(\tau_{AB}(\pi_1)) \cdot f(\phi_A(\pi_1)) \cdot g(\phi_B(\pi_1)) > 0, \text{ and}$$
$$h(\pi_2) = m(\tau_{AB}(\pi_2)) \cdot f(\phi_A(\pi_2)) \cdot g(\phi_B(\pi_2)) > 0.$$

Since $\pi_1$ and $\pi_2$ disagree on the relative ranking of items in $A$, this factorization implies in particular that both $f(\phi_A = i|j) > 0$ and $f(\phi_A = j|i) > 0$. Since $h(\pi_3) > 0$, it must also be that each of $m(\tau_{AB}(\pi_3))$, $f(\phi_A(\pi_3))$, and $g(\phi_B(\pi_3))$ have positive probability. We can therefore swap the relative ranking of $A$, $\phi_A$, to obtain a new ranking which has positive probability since all of the terms in the decomposition of this new ranking have positive probability. $\qquad\square$

Lemma 29 below provides conditions under which removing a vertical bar from one of the rankings in $X$ will not change the support of a completely riffle independent distribution. To illustrate with an example, consider a completely decomposable observation $\mathcal{O}$ which contains the partial ranking $S_\gamma \pi = Corn, Peas | Apples, Oranges$ as a subset. What Lemma 29 guarantees is that, if, in addition, there exists any element $\tilde{\pi}$ in $\mathcal{O}$ which *disagrees* with $S_\gamma \pi$ on the relative ordering of, say, *Peas* and *Oranges*, then in fact the partial ranking $S_{\gamma'} \pi' \subset = Corn, Peas, Apples, Oranges$ (with the bar removed from $S_\gamma \pi$) must also be a subset of $\mathcal{O}$. Formally,

**Lemma 29.** *Let $S_\gamma \pi = \Omega_1 | \ldots | \Omega_i | \Omega_{i+1} | \ldots | \Omega_k$ be a partial ranking on item set $\Omega$, and $S_{\gamma'} \pi' = \Omega_1 | \ldots | \Omega_i \cup \Omega_{i+1} | \ldots | \Omega_k$, the partial ranking in which the sets $\Omega_i$ and $\Omega_{i+1}$ are merged. Let $a_1 \in \cup_{j=1}^{i} \Omega_j$ and $a_2 \in \cup_{j=i+1}^{k} \Omega_j$. If $\mathcal{O}$ is any element of $\mathfrak{C}$ such that $S_\gamma \pi \subset \mathcal{O}$ and there additionally exists a ranking $\tilde{\pi} \in \mathcal{O}$ which disagrees with $S_\gamma \pi$ on the relative ordering of $a_1, a_2$, then $S_{\gamma'} \pi' \subset \mathcal{O}$.*

*Proof.* The key strategy in our proof of Lemma 29 is to argue that large subsets of rankings must be contained in a completely decomposable observation $\mathcal{O}$ by decomposing rankings into transpositions and invoking the technical lemma from above (Lemma 28) repeatedly. See the Appendix for details. $\qquad\square$

We now can use Lemma 29 to show that the reverse inclusion of Proposition 26 also holds, establishing that the two sets $\textsc{rspan}(X)$ and $\textsc{pspan}(X)$ are in fact equal and thereby proving the desired result, that $\mathfrak{C} \subset \mathcal{P}$.

*Proof. (of Proposition 26)* At each iteration $t$, Algorithm 2 produces a set of partial rankings, $X_t$. We denote the union of all partial rankings at time $t$ as $\tilde{X}_t \equiv \bigcup_{S_\gamma \sigma \in X_t} S_\gamma \sigma$. Note that $\tilde{X}_0 = X$ and $\tilde{X}_T = \textsc{pspan}(X)$. The idea of our proof will be to show that at each iteration $t$, the following set inclusion holds: $\textsc{rspan}(\tilde{X}_t) \subset \textsc{rspan}(\tilde{X}_{t-1})$. If indeed this holds, then after the final iteration $T$, we will have shown that:

$$\text{PSPAN}(X) = \tilde{X}_T, \qquad \text{(Proposition 27)}$$
$$\subset \text{RSPAN}(\tilde{X}_T), \qquad \text{(Monotonicity, Proposition 24)}$$
$$\subset \text{RSPAN}(\tilde{X}_0), \qquad \text{(since } \text{RSPAN}(\tilde{X}_t) \subset \text{RSPAN}(\tilde{X}_{t-1}), \text{ shown below),}$$
$$\subset \text{RSPAN}(X) \qquad (\tilde{X}_0 = X, \text{ see Algorithm 2})$$

which would prove the Proposition.

It remains now to show that $\text{RSPAN}(\tilde{X}_t) \subset \text{RSPAN}(\tilde{X}_{t-1})$. We claim that $\tilde{X}_t \subset \text{RSPAN}(\tilde{X}_{t-1})$. Let $\sigma \in \tilde{X}_t$. If $\sigma \in \tilde{X}_{t-1}$, then since $\tilde{X}_{t-1} \subset \text{RSPAN}(\tilde{X}_{t-1})$, we have $\sigma \in \text{RSPAN}(\tilde{X}_{t-1})$ and the proof is done. Otherwise, $\sigma \in \tilde{X}_t \backslash \tilde{X}_{t-1}$. In this second case, we use the fact that at iteration $t$, the vertical bar between $\Omega_i$ and $\Omega_{i+1}$ was deleted from the partial ranking $S_\gamma \pi = \Omega_1 | \dots | \Omega_i | \Omega_{i+1} | \dots | \Omega_k$ (which is a subset of $\tilde{X}_{t-1}$) to form the partial ranking $S_{\gamma'} \pi' = \Omega_1 | \dots | \Omega_i \cup \Omega_{i+1} | \dots | \Omega_k$. (which is a subset of $\tilde{X}_t$). Furthermore, in order for the vertical bar to have been deleted by the algorithm, there must have existed some partial ranking (and therefore some full ranking $\omega'$) that disagreed with $S_\gamma \pi$ on the relative ordering of items $a_1, a_2$ on opposite sides of the bar. Since $\sigma \in \tilde{X}_t \backslash \tilde{X}_{t-1}$ we can assume that $\sigma \in S_{\gamma'} \pi'$.

We now would like to apply Lemma 29. Note that for any $\mathcal{O} \in \mathfrak{C}$ such that $\tilde{X}_{t-1} \subset \mathcal{O}$, we also have $S_\gamma \pi \subset \mathcal{O}$, since $S_\gamma \pi \subset \tilde{X}_{t-1}$. An application of Lemma 29 then shows that $S_{\gamma'} \pi' \subset \mathcal{O}$ and therefore that $\sigma \in \mathcal{O}$.

We have shown in fact that $\sigma \in \mathcal{O}$ holds for *any* observation $\mathcal{O} \in \mathfrak{C}$ such that $\tilde{X}_{t-1} \subset \mathcal{O}$, and therefore taking the intersection of supports over all $\mathcal{O} \in \mathfrak{C}$, we see that $\tilde{X}_t \subset \text{RSPAN}(\tilde{X}_{t-1})$. Taking the RSPAN of both sides yields:

$$\text{RSPAN}(\tilde{X}_t) \subset \text{RSPAN}(\text{RSPAN}(\tilde{X}_{t-1})), \qquad \text{(Subset preservation, Proposition 24)}$$
$$\subset \text{RSPAN}(\tilde{X}_{t-1}). \qquad \text{(Idempotence, Proposition 24)}$$

$\qquad\qquad\qquad\qquad\qquad\qquad\qquad\qquad\qquad\qquad\qquad\qquad\qquad\qquad\qquad\qquad\qquad\qquad$ $\square$

## 5.3 Going Beyond Subset Observations

Though we have stated all of our results so far for subset observations, we now comment on what our theory would look like if we had considered general likelihood functions. In order to avoid confusion, we here refer to a more general class of functions that we call *completely decomposable functions*, instead of the completely decomposable subset observations of Definition 11.

**Definition 30.** A function $h : S_n \to \mathbb{R}$ is called a *completely decomposable function* if it factors riffle independently with respect to every hierarchy over the item set $\Omega$. We denote the collection of all possible completely decomposable functions as $\widetilde{\mathfrak{C}}$.

As we discuss, $\mathfrak{C}$ and $\widetilde{\mathfrak{C}}$ are very nearly the same. It is quite simple to restate Theorem 19 with respect to the general case of completely decomposable functions:

**Theorem.** *Every partial ranking indicator function is a completely decomposable function.*

Unfortunately, the proof of its converse (Theorem 21) does not easily generalize, and instead can only be used to show that the *support* ($\{\sigma \in S_n : h(\sigma) > 0\}$) of every completely decomposable function is a partial ranking. It is natural, however, to suspect that a full converse does indeed exist — that every completely decomposable function is proportional to the indicator function of some partial ranking. In fact, this suspected converse only *almost* holds. We have:

**Theorem.** *If $h$ is any completely decomposable function supported on a partial ranking $S_\gamma \pi = \Omega_1 | \ldots | \Omega_r$ where $|\Omega_i| \neq 2$ for all $i = 1, \ldots, r$, then $h$ is proportional to the indicator function on $S_\gamma \pi$.*

*Proof.* See Appendix. □

**Example 31.** *For completely decomposable functions, it is not possible to do away with the assumption that $|\Omega_i| \neq 2$ for all $i$. As an example, the function defined below as:*

$$h(\sigma) = \begin{cases} 2/3 & \text{if } \sigma = Corn|Peas|Apples \\ 1/3 & \text{if } \sigma = Peas|Corn|Apples \\ 0 & \text{otherwise} \end{cases},$$

*is supported on the partial ranking $S_\gamma \pi = Corn, Peas|Apples$ (where $|\Omega_1| = 2$), and is not proportional to any indicator function (i.e., it is not uniform on rankings which are not assigned positive probability).*

*However, it is still possible to show that $h$ is a completely decomposable function. To prove so, it is necessary to establish only three things: that $\{Corn, Peas\}$ and $\{Apples\}$ are riffle independent, that $\{Corn, Apples\}$ and $\{Peas\}$ are riffle independent, and that $\{Peas, Apples\}$ and $\{Corn\}$ are riffle independent. For example, with respect to the partitioning into sets $A = \{Corn, Apples\}$ and $B = \{Peas\}$, we see that*

$$h(\sigma) = m(\tau_{AB}(\sigma)) \cdot f(\phi_A(\sigma)) \cdot g(\phi_B(\sigma)),$$

*where:*

$$m(\tau_{AB}) = \begin{cases} 2/3 & \text{if } \tau_{AB} = A|B|A \\ 1/3 & \text{if } \tau_{AB} = B|A|A \\ 0 & \text{if } \tau_{AB} = A|A|B \end{cases}, \quad f(\sigma_A) = \begin{cases} 1 & \text{if } \sigma_{\{AC\}} = A|C \\ 0 & \text{otherwise} \end{cases}, \quad g(\sigma_B) = 1.$$

*Therefore, when $|\Omega_i| = 2$, it is possible to have completely decomposable functions which are not uniform on their supports.*

### 5.4 Conditioning on Noisy Observations

We conclude this section with a remark on handling noise in observations. While we have assumed in this paper that observed partial rankings are always consistent with a user's underlying full ranking, there are situations in which one may wish to model a noisier setting, where the partial rankings may be misreported with some small probability. A natural model that accounts for noise, for example, might be:

$$L(\mathcal{O}|\sigma) = \begin{cases} 1 - \varepsilon & \text{if } \sigma \in \mathcal{O} \\ \frac{\varepsilon}{|\mathcal{O}| - 1} & \text{otherwise} \end{cases}. \tag{5.2}$$

If a prior distribution factorizes with respect to a hierarchy $H$, then conditioning on the noisy likelihood of Equation 5.2 results in a posterior distribution which can be written as a weighted mixture of the prior distribution and the posterior that would have resulted from conditioning on a noise-free observation. While each component of this posterior distribution factorizes with respect to $H$, the mixture itself does not factor in general (and should not factor according to our theory). As a result, iteratively conditioning on multiple partial rankings according to the noisy likelihood function above would quickly lead to an unmanageable number of mixture components. We therefore believe that approximate inference methods for conditioning on multiple noisy partial ranking observations is a fruitful area for further research.

## 6. Model Estimation from Partially Ranked Data

In many ranking based applications, datasets are predominantly composed of partial rankings rather than full rankings due to the fact that for humans, partial rankings are typically easier and faster to specify. In addition, many datasets are heterogeneous, containing partial ranking of different types. For example, in the American Psychological Association as well as the Irish House of Parliament elections, voters are allowed to specify their top-$k$ candidate choices for any value of $k$ (see Figures 7(a) and 7(b)). In this section we use the efficient inference algorithm proposed in Section 5 for estimating a riffle independent model from partially ranked data. Because estimating a model using partially ranked data is typically considered to be more difficult than estimating one using only full rankings, a common practice (e.g., see Huang & Guestrin, 2010) has been to simply ignore the partial rankings in a dataset. The ability of a method to incorporate *all* of the available data however, can lead to significantly improved model accuracy as well as wider applicability of that method. In this section, *we propose the first efficient method for estimating the structure and parameters of a hierarchical riffle independent model from heterogeneous datasets consisting of arbitrary partial ranking types.* Central to our approach is the idea that given someone's partial preferences, we can use the efficient algorithms developed in the previous section to infer his full preferences and consequently apply previously proposed algorithms which are designed to work with full rankings.

### 6.1 Censoring Interpretations of Partial Rankings

The model estimation problem for full rankings is stated as follows. Given i.i.d. training examples $\sigma^{(1)}, \dots, \sigma^{(m)}$ (consisting of full rankings) drawn from a hierarchical riffle independent distribution $h$, recover the structure and parameters of $h$.

In the partial ranking setting, we again assume i.i.d. draws, but that each training example $\sigma^{(i)}$ undergoes a censoring process producing a partial ranking consistent with $\sigma^{(i)}$. For example, censoring might only allow for the ranking of the top-$k$ items of $\sigma^{(i)}$ to be observed. While we allow for arbitrary types of partial rankings to arise via censoring, we make a common assumption that the partial ranking type resulting from censoring $\sigma^{(i)}$ does not depend on $\sigma^{(i)}$ itself.

## 6.2 Algorithm

We treat the model estimation from partial rankings problem as a missing data problem. As with many such problems, if we could determine the full ranking corresponding to each observation in the data, then we could apply algorithms which work in the completely observed data setting. Since full rankings are not given, we utilize an Expectation-Maximization (EM) approach in which we use inference to compute a posterior distribution over full rankings given the observed partial ranking. In our case, we then apply the algorithms from Huang and Guestrin (2010, 2012) which were designed to estimate the hierarchical structure of a model and its parameters from a dataset of full rankings.

Given an initial model $h$ and a collection of training examples $\{\mathcal{O}^{(1)}, \mathcal{O}^{(2)}, \ldots, \mathcal{O}^{(m)}\}$ consisting of partial rankings, our EM-based approach alternates between the following two steps until convergence is achieved.

- **(E-step)**: For each observation, $\mathcal{O}^{(i)} = S_{\gamma^{(i)}} \pi^{(i)}$, in the training examples, we use inference to compute a posterior distribution over the full ranking $\sigma$ that could have generated $\mathcal{O}^{(i)}$ via censoring, $h(\sigma | \mathcal{O}^{(i)} = S_{\gamma^{(i)}} \pi^{(i)})$. Since the observations take the form of partial rankings and are hence completely decomposable, we use the efficient algorithms in Section 5 to perform the E-step.

- **(M-step)**: In the M-step, one maximizes the expected log-likelihood of the training data with respect to the model. When the hierarchical structure of the model has been provided, or is known beforehand, our M-step can be performed using standard methods for optimizing parameters. When the structure is *unknown*, we use a *structural EM* approach, which is analogous to methods from the graphical models literature for structure learning from incomplete data (Friedman, 1997, 1998).

  Unfortunately, the (riffled independence) structure learning algorithm of Huang and Guestrin (2010) is unable to directly use the posterior distributions computed from the E-step. Instead, observing that sampling from riffle independent models can be done efficiently and exactly (as opposed to, for example, MCMC methods), we simply sample full rankings from the posterior distributions computed in the E-step and pass these full rankings into the structure learning algorithm of Huang and Guestrin (2010). The number of samples that are necessary, instead of scaling factorially, scales according to the number of samples required to detect riffled independence (which under mild assumptions is polynomial in $n$, Huang & Guestrin, 2010).

## 7. Related Work

Rankings and permutations have recently become an active area of research in machine learning due in part to the hinge role that they play in information retrieval and preference elicitation. Algorithms such as the RankSVM (Joachims, 2002) and RankBoost (Freund, Iyer, Schapire, & Singer, 2003), for example, have been successful in the large scale ranking problems that appear in web search. The main aims of our work differ from these web scale settings however — instead of seeking a single 'optimal' ranking with respect to some objective function, we seek an understanding of a large collection of rankings via density estimation. In the following, we outline two major lines of research which have influenced our work.

### 7.1 Additive and Multiplicative Decompositions

Our paper builds in particular upon a thread of recent work on tractable models for permutation data based on function decompositions. Kondor, Howard, and Jebara (2007) and Huang, Guestrin, and Guibas (2008, 2009) considered additive decompositions of a distribution into a weighted sum of Fourier basis functions. These papers show that low-frequency Fourier assumptions can often be effective for coping with the representational complexity of working with distributions over permutations. They show in particular that conditioning prior distributions on the 'low frequency' likelihood functions that often arise in multiobject tracking problems can be performed especially efficiently.

Unfortunately, low frequency assumptions are not as applicable for distributions defined over rankings, and to address ranking problems specifically, Huang and Guestrin (2009, 2010) introduced the concept of riffled independence as a useful generalization of probabilistic independence for rankings. Using *multiplicative decompositions* based on riffled independence, we showed that it is possible to learn the hierarchical structure of a model given a fully ranked dataset. While our previous papers on the topic of riffled independence focused more on problems related to efficiently representing distributions, the main focus of our current paper lies in efficient reasoning/inference and tackling human task complexity by considering partial rankings.

It is interesting to note that while it is natural and efficient to condition a Fourier based representation on low-frequency observations (involving a very small number of items) such as $\mathcal{O} =$"Alice is in third place", a multiplicative decomposition based on riffled independence would not be able to efficiently condition on the same observation. On the other hand, multiplicative decompositions allow us to condition on top-$k$ observations efficiently (independently of the size of $k$), whereas top-$k$ observations would be difficult to handle in a Fourier theoretic setting (except for very small $k$).

### 7.2 Mallows Models

Our work also fits into a larger body of research about the well known Mallows distribution over rankings, parameterized by:

$$h(\sigma; \phi, \sigma_0) \propto \phi^{-d_\tau(\sigma, \sigma_0)}, \tag{7.1}$$

where the function $d_\tau$ refers to the *Kendall's tau* distance metric on rankings. A Mallows distribution (Equation 7.1) can always be shown to be a special case of a hierarchical riffle independent model in which items are sequentially factored out of the model one by one (Huang, 2011) (see Figure 6).

Mallows models (as well as other similar distance based models) have the advantage that they can compactly represent distributions for very large $n$, and admit conjugate prior distributions (Meila, Phadnis, Patterson, & Bilmes, 2007). Estimating parameters has been a popular problem for statisticians — recovering the optimal $\sigma_0$ from data is known as the *consensus ranking* or *rank aggregation* problem and is known to be $NP$-hard (Bartholdi, Tovey, & Trick, 1989). Many authors have focused on approximation algorithms instead.

Like Gaussian distributions, Mallows models tend to lack flexibility, and so Lebanon and Mao (2008) propose a nonparametric model of ranked (and partially ranked) data based on placing weighted Mallows kernels on top of training examples, which, as they show, can realize a far richer class of distributions, and can be learned efficiently. However, they do not address the inference problem, and it is not immediately clear in many Mallows models papers whether one can efficiently perform inference operations like marginalization and conditioning in such models. Riffle independent models, on the other hand, encompass a class of distributions which is both rich as well as interpretable, and additionally, we have identified precise conditions under which efficient conditioning is possible (the conditions being that the observations take the form of partial rankings).

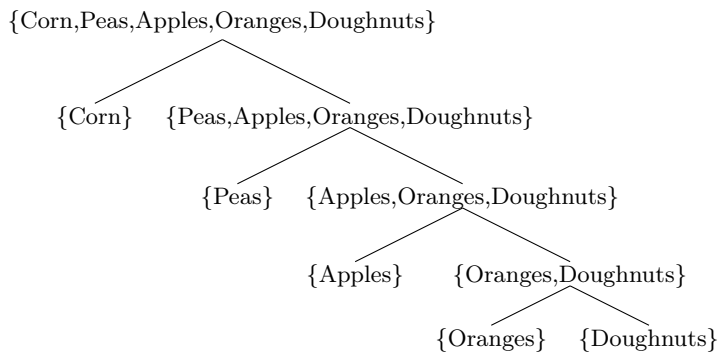

Figure 6: A Mallows model always factors according to what we refer to as a 'chain' structure in which items are factored out one by one. The Mallows distribution over five items from our food item set with mode (or central ranking) at $\sigma_0 = Corn|Peas|Apples|Oranges|Doughnuts$, for example, must factor according to the above hierarchical structure.

There are several recent works to model partial rankings using Mallows based models. Busse, Orbanz, and Buhmann (2007) learned finite mixtures of Mallows models from top-$k$ data (also using an EM approach). Lebanon and Mao (2008), as we have mentioned, developed a nonparametric model based on Mallows models which can handle arbitrary types of partial rankings. In both settings, a central problem is to marginalize a Mallows model over all full rankings which are consistent with a particular partial ranking. To do so efficiently, both papers rely on the fact (first shown in Fligner & Verducci, 1986) that this marginalization step can be performed in closed form. This closed form equation of Fligner and Verducci (1986), however, can be seen as a very special case of our setting since Mallows models can always be shown to factor riffle independently according to a chain structure. Specifically, to compute the sum over rankings which are consistent with a partial ranking $S_\gamma\sigma$, it is necessary to condition on $S_\gamma\sigma$, and to compute the normalization constant of the resulting function. The conditioning step can be performed using the methods that we have described in this paper, and the normalization constant can be computed by multiplying the normalization constant of each factor of the hierarchical decomposition. Thus, instead of resorting to the more complicated mathematics of inversion combinatorics, our theory of complete decomposability offers a simple conceptual way to understand why Mallows models can be conditioned efficiently on partial ranking observations.

Finally in recent related work, Lu and Boutilier (2011) considered an even more general class of observations based on DAG (directed acyclic graph) based observations in which probabilities of rankings which are not consistent with a DAG of relative ranking relations are set to zero. Lu and Boutilier show in particular that the conditioning problem for their DAG-based class of observations is $\#P$-hard. They additionally propose an efficient rejection sampling method for performing probabilistic inference within the general class of DAG observations and prove that the sampling method is *exact* for the class of partial rankings that we have discussed in this paper.

## 8. Experiments

In this section, we demonstrate our method for learning hierarchical riffle independent models from partial rankings on simulated data as well as real datasets taken from different domains. In all experiments, we initialize distributions to be uniform, and do not use random restarts.

### 8.1 Datasets

In addition to roughly 5000 full rankings, the APA dataset has over 10,000 top-$k$ rankings of 5 candidates. In previous work, we had used only the full rankings of the APA data (Huang & Guestrin, 2010, 2012), but now we are able to use the entire dataset. Figure 7(a) plots, for each $k \in \{1, \dots, 5\}$, the number of ballots in the APA data of length $k$.

Likewise, the *Meath dataset* (Gormley & Murphy, 2007) which was taken from the 2002 Irish Parliament election has over 60,000 top-$k$ rankings of 14 candidates. As with the APA data, we had used only the full rankings of the Meath data in previous work, but here we use the entire dataset. Figure 7(b) plots, for each $k \in \{1, \dots, 14\}$, the number of ballots in the Meath data of length $k$. In particular, note that the vast majority of ballots in the dataset consist of partial rather than full rankings, with over half of the electorate preferring to list only their favorite three or four candidates. We can run inference (Algorithm 1) on over 5000 top-$k$ examples for the Meath data in 10 seconds on a dual 3.0 GHz Pentium machine with an unoptimized Python implementation. Using 'brute force' inference, we estimate that the same job would require roughly one hundred years.

We extracted a third dataset from a database of *searchtrails* collected by White and Drucker (2007), in which browsing sessions of roughly 2000 users were logged during 2008-2009. In many cases, users are unlikely to read articles about the same news story twice, and so it is often possible to think of the order in which a user reads through a collection of articles as a top-$k$ ranking over articles concerning a particular story/topic. The ability to model visit orderings would allow us to make long term predictions about user browsing behavior, or even recommend 'curriculums' over articles for users. We ran our algorithms on roughly 300 visit orderings for the eight most popular posts from `www.huffingtonpost.com` concerning 'Sarah Palin', a popular subject during the 2008 U.S. presidential election. Since no user visited every article, there are no full rankings in the data and thus there does not even exist the option of learning using only the subset of full rankings.

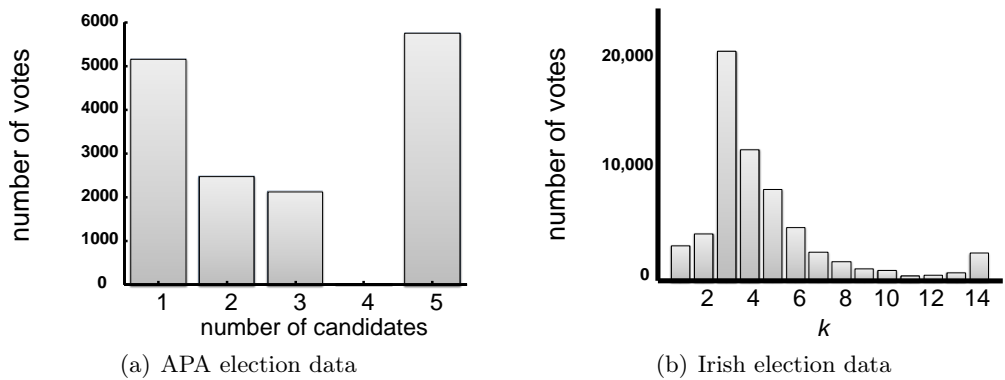

(a) APA election data

(b) Irish election data

Figure 7: Histograms of top-$k$ ballot lengths in the APA and Irish election datasets. Whereas the majority of the electorate provided full rankings in the APA election data (probably due to the fact that there were only five candidates), the vast majority of voters in the Irish election data provided only their top-3 or top-4 choices.

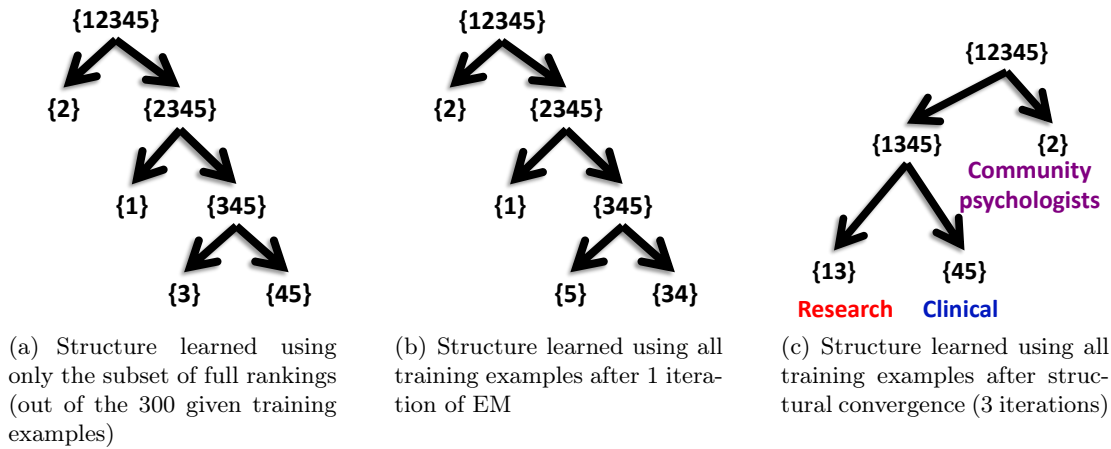

(a) Structure learned using only the subset of full rankings (out of the 300 given training examples)

(b) Structure learned using all training examples after 1 iteration of EM

(c) Structure learned using all training examples after structural convergence (3 iterations)

Figure 8: Structure learning with a subset of the APA dataset (300 rankings, randomly sampled, including both full and partial rankings).

## 8.2 APA Structure Learning Results

Due to the unordinarily large number of full rankings in the APA data, the gains made by additionally using partially ranked data are insignificant. To better illustrate the benefits of partial rankings, we subsampled a dataset of 300 rankings (including both full and partial rankings) and present results with this smaller dataset. Performing structure learning using *only* the full rankings of these 300 training examples (consisting of roughly 100 examples), one obtains the structure in Figure 8(a), which can be seen to not match the 'correct' structure of Figure 3(a) which was learned using 5000 full rankings. Figures 8(b) and 8(c)

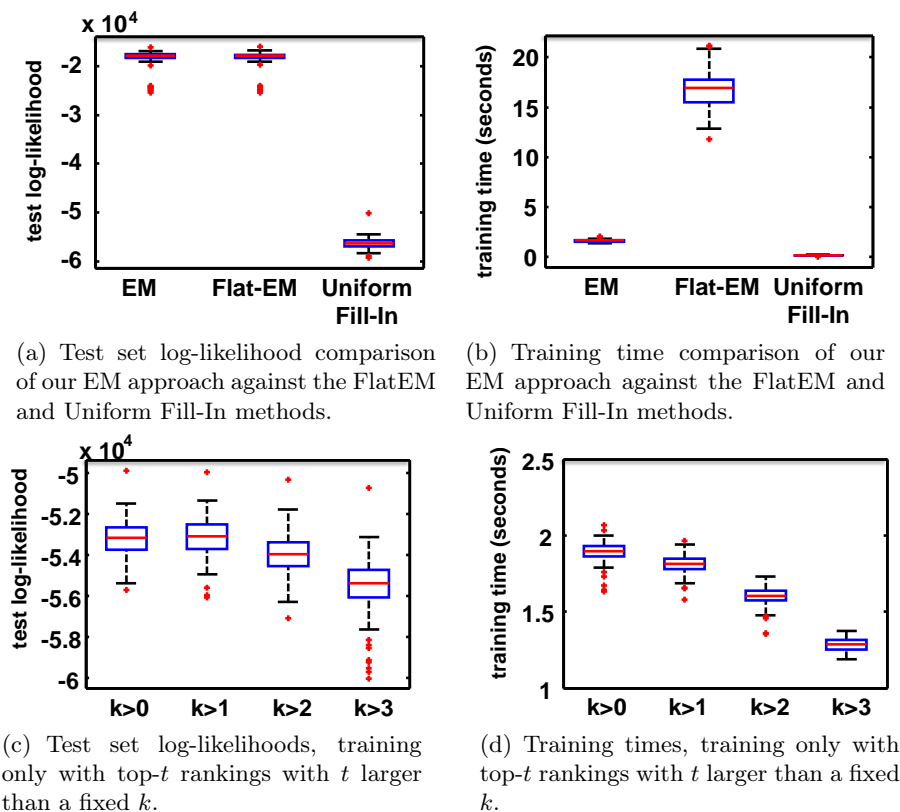

(a) Test set log-likelihood comparison of our EM approach against the FlatEM and Uniform Fill-In methods.

(b) Training time comparison of our EM approach against the FlatEM and Uniform Fill-In methods.

(c) Test set log-likelihoods, training only with top-$t$ rankings with $t$ larger than a fixed $k$.

(d) Training times, training only with top-$t$ rankings with $t$ larger than a fixed $k$.

Figure 9: APA experimental results — each experiment repeated with 200 bootstrapped resamplings of the data

plot the results of our EM algorithm with the former displaying the resulting structure after just a single EM iteration and the latter the result after structural convergence, which occurs by the third iteration, showing that our method can learn the 'correct' structure given just 300 training examples.

We compared our EM algorithm against two alternative baseline approaches that we refer to in our plots as *FlatEM* and *Uniform Fill-in*. The FlatEM algorithm is the same as the EM algorithm above except for two details: (1) it performs conditioning exhaustively instead of exploiting the factorized model structure, and (2) it performs the M-step without sampling. The Uniform Fill-in approach treats every top-$k$ ranking in the training set as a uniform collection of votes for all of the full rankings consistent with that top-$k$ ranking, and is accomplished by using just one iteration of our EM algorithm.

In Figure 9(a) we plot test set loglikelihoods corresponding to each approach, with EM and FlatEM having almost identical results and both performing much better than the Uniform Fill-in approach. On the other hand, Figure 9(b), which compares running times of the three approaches, shows that FlatEM can be far more costly (for most datasets, it cannot even be run in a reasonable amount of time).

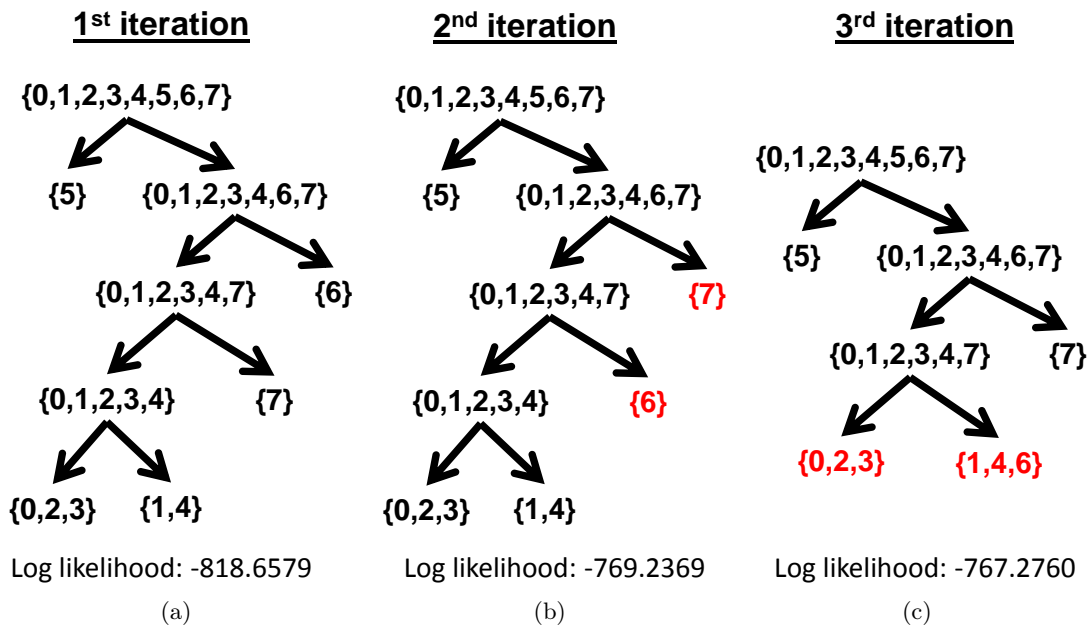

Figure 10: Iterations of Structure EM for the Sarah Palin data with structural changes at each iteration highlighted in red. Structural convergence occurs after just three iterations. Note that this structure was discovered using only visit orders, and that no text information from pages was incorporated in the learning process. This figure is best viewed in color.

To verify that partial rankings *do* indeed make a difference in the APA data, we plot the results of estimating a model from the subsets of APA training data consisting of top-$k$ rankings with length larger than some fixed $k$. Figures 9(c) and 9(d) show the log-likelihood and running times for $k = 0, 1, 2, 3$ with $k = 0$ being the entire training set and $k = 3$ being the subset of training data consisting only of full rankings. As our results show, including partial rankings does indeed help on average for improving test log-likelihood (with diminishing returns).

## 8.3 Structure Discovery with EM with Larger $n$.

Our experiments have led to several observations about using EM for learning with partial rankings. First, we observe that typical runs converge to a fixed structure quickly, with no more than three EM iterations. Figure 10 shows the progress of EM on the Sarah Palin data, whose structure converges by the third iteration. As expected, the log-likelihood increases at each iteration, and we remark that the structure becomes more interpretable — for example, the leaf set $\{0, 2, 3\}$ corresponds to the three posts about Palin's wardrobe before the election, while the posts from the leaf set $\{1, 4, 6\}$ were related to verbal gaffes made by Palin during the campaign. Notice that this structure is discovered purely using data about visit orders and that no text information was used in our experiments.

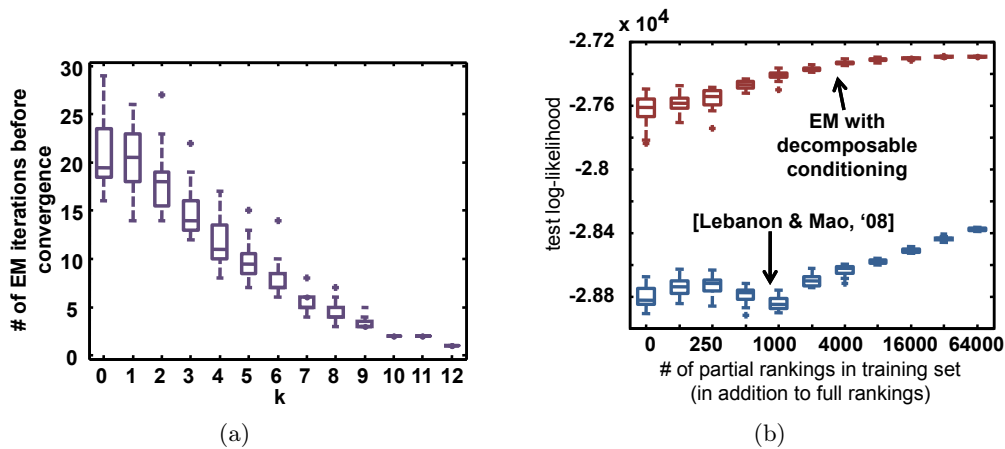

(a)                                  (b)

Figure 11: (a): Number of EM iterations required for convergence if the training set only contains rankings of length longer than $k$. (b): Density estimation from synthetic data. We plot test loglikelihood when learning from 343 full rankings and between 0 and 64,000 additional partial rankings.

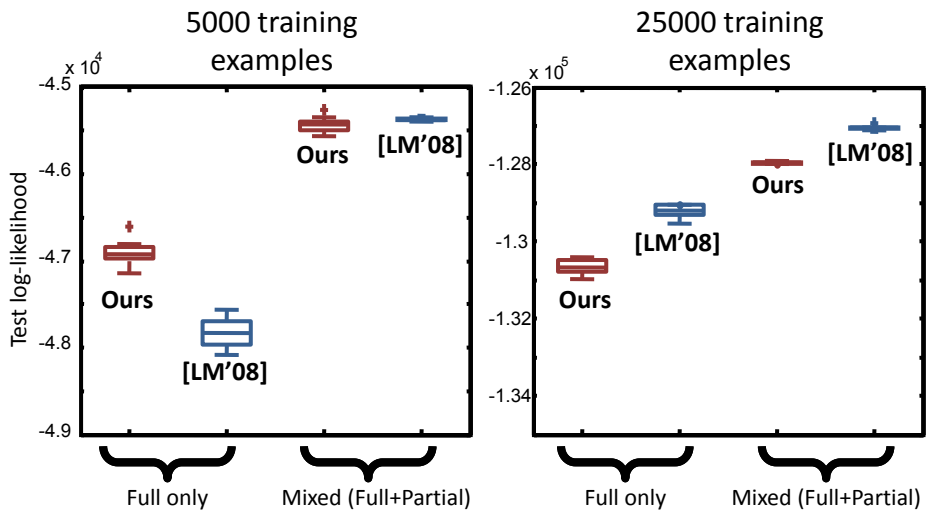

Figure 12: Density estimation from small (5000 examples) and large subsets (25000 examples) of the Meath data. We compare our method against the work by Lebanon and Mao (2008) in two settings: (1) training on all available data and (2) training on the subset of full rankings.

Secondly, the number of EM iterations required to reach convergence in log-likelihood depends on the types of partial rankings observed. We ran our algorithm on subsets of the Meath dataset, each time training on $m = 2000$ rankings all with length larger than some fixed $k$. Figure 11(a) shows the number of iterations required for convergence as a function of $k$ (with 20 bootstrap trials for each $k$). We observe fastest convergence for datasets consisting of almost-full rankings and slowest convergence for those consisting of almost-empty rankings, with almost 25 iterations necessary if one trains using rankings of all types. Finally we remark that the model obtained after the first iteration of EM is interesting and can be thought of as the result of pretending that each voter is completely ambivalent regarding the $n - k$ unspecified candidates.

## 8.4 The Value of Partial Rankings

We now verify again with larger $n$ that using partial rankings in addition to full rankings allows us to achieve better density estimates. We first learned models from synthetic data drawn from a hierarchy, training using 343 full rankings plus varying numbers of partial ranking examples (ranging between 0-64,000). We repeat each setting with 20 bootstrap trials, and for evaluation, we compute the log-likelihood of a testset with 5000 examples. For speed, we learn a structure $H$ only once and fix $H$ to learn parameters for each trial. Figure 11(b), which plots the test log-likelihood as a function of the number of partial rankings made available to the training set, shows that we are indeed able to learn more accurate distributions as more and more data in the form of partial rankings are made available.

## 8.5 Comparing to a Nonparametric Model

Comparing the performance of riffle independent models to other approaches was not possible in previous work since we had not been able to handle partial rankings. Using the methods developed in our current paper, however, we compare riffle independent models with the state-of-the-art nonparametric estimator of Lebanon and Mao (2008) (to which we hereby refer as the LM08 estimator) on the same data (setting their regularization parameter to be $C = 1,2,5,$ or 10 via a validation set). Figure 11(b) shows (naturally) that when the data are drawn synthetically from a riffle independent model, then our EM method significantly outperforms the LM08 estimator. We remark that in theory, the LM08 is guaranteed to catch up in performance (under appropriate conditions) given enough training examples.

For the Meath data, which is only approximately riffle independent, we trained on subsets of size 5,000 and 25,000 (testing on remaining data). For each subset, we evaluated our EM algorithm for learning a riffle independent model against the LM08 estimator when (1) using only full ranking data, and (2) using all data. As before, both methods do better when partial rankings are made available.

For the smaller training set, the riffle independent model performs as well or better than the LM'08 estimator. For the larger training set of 25,000, we see that the nonparametric method starts to perform slightly better on average, the advantage of a nonparametric model being that it is guaranteed to be consistent, converging to the correct model given enough data. The advantage of riffle independent models, however, is that they are simple, interpretable, and can highlight global structures hidden within the data.

## 9. Future Directions

There remain several possible extensions to the current work. We list a few such open questions and extensions in the following.

### 9.1 Inference with Incomplete Rankings

We have shown in this paper that one can exploit riffled independence structure to condition on an observation if and only if it takes the form of a partial ranking. While the space of partial rankings is both rich and useful in many settings, it does not cover an important class of observations: that of *incomplete rankings*, which are defined to be a ranking (or partial ranking) of a subset of the itemset $\Omega$. For example, Theorem 21 shows that the conditioning problem for pairwise observations of the form "Apples are preferred over Bananas" is nondecomposable. Note that top-$k$ rankings are considered to be "complete" rankings since they implicitly rank all other items in the last $n - k$ positions.

How then, can we tractably condition on incomplete rankings? One possible approach is to convert to a Fourier representation using the methods from (Huang & Guestrin, 2012), then conditioning on a pairwise ranking observation using the Fourier domain conditioning algorithm proposed in (Huang et al., 2008). This Fourier domain approach would be useful if one were particularly interested in low-order marginal probabilities of the posterior distributions.

When the Fourier approach is not viable, another option may be to assume that the posterior distribution takes on a particular riffle independent structure (in the same way that mean field methods from the graphical models literature would assume a factorized posterior). The research question of interest is: which hierarchical structure should be used for the purposes of approximating the posterior?

### 9.2 Reexamining Data Independence Assumptions

In this paper, we have assumed throughout that training examples are independent and identically distributed. However in practice these are not always safe assumptions as a number of factors can impact the validity of both. For example, in an internet survey in which a user must perform a series of preference ranking tasks in sequence, a concern is that the user's prior ranking tasks may bias the results of his future rankings.

Another source of bias lies in the reference ranking that may be displayed, in which the user is asked to rearrange items by 'dragging and dropping'. On the one hand, showing everyone the same reference ranking may bias the resulting data. But on the other hand, showing every user a different reference ranking may mean that the training examples are not exactly identically distributed.

Yet another form of bias lies in the partial ranking types that are reported in data. To formulate our EM algorithm, we have assumed that a user's preferences does not influence whether he chooses to, say, report a full ranking instead of a top-3 ranking. In practice, however, partial ranking types and user preferences are often correlated. In the Irish elections, for example, where there is typically only one Sinn Fein candidate, those who rank Sinn Fein first are typically more likely to have only reported their top-1 choice.

Understanding, identifying, and finally, learning in spite of the different types of biases that may occur in eliciting preference data remains a fundamental problem in ranking.

## 9.3 Probabilistic Modeling of Strategic Voting

It is interesting to consider the differences between the actual vote distributions considered in this paper against the approximate riffle independent distributions. Take the APA dataset, for example, in which the optimal approximation by a riffle independent hierarchy reflects the underlying political coalitions within the organization. Upon comparison between the approximation and the empirical distribution, however, some marked differences arise. For example, the riffle independent approximation underestimates the number of votes obtained by candidate 3 (a research psychologist) who ultimately won the election.

One possible explanation for the discrepancy may lie in the idea that voters tend to vote strategically in APA elections, placing stronger candidates of opposing political coalitions lower in the ranking, rather than revealing their true preferences. An interesting line of future work lies in detecting and studying the presence of such strategic voting in election datasets. Open questions include (1) verifying mathematically whether strategic voting does indeed exist in, say, the APA election data, and (2) if so, why the strategic voting effect is not strong enough to overwhelm our riffled independence structure learning algorithms, and (3) how strategic voting can manifest itself in partial ranking votes.

## 10. Conclusion

In probabilistic reasoning problems, it is often the case that certain data types suggest certain distribution representations. For example, sparse dependency structure in the data often suggests a Markov random field (or other graphical model) representation (Friedman, 1997, 1998). For low-order permutation observations (depending on only a few items at a time), recent work (Huang et al., 2009; Kondor, 2008) has shown that a Fourier domain representation is appropriate. For preference ranking scenarios, one must contend with 'human task complexity' — the difficulty involved for a human to rank a long list of items and often leads to partially, instead of fully ranked data. In this paper, we have shown that when data takes the form of partial rankings, then hierarchical riffle independent models are a natural representation.

As with conjugate priors, we showed that a riffle independent model is guaranteed to retain its factorization structure after conditioning on a partial ranking (which can be performed in efficiently). Most surprisingly, our work shows that observations which do not take the form of partial rankings are not amenable to simple multiplicative update based conditioning algorithms. Finally, we showed that it is possible to learn hierarchical riffle independent models from partially ranked data, significantly extending the applicability of previous work.

## Acknowledgments

This project was formulated and largely conducted during an internship by Jonathan Huang at Microsoft Research. Additional work was supported in part by ONR under MURI N000140710747, and ARO under MURI W911NF0810242. Carlos Guestrin was funded in part by NSF Career IIS-064422. We thank Eric Horvitz, Ryen White, Dan Liebling, and Yi Mao for discussions.

## Appendix A. Proofs

In this appendix, we provide supplementary proofs of some of the theoretical results in this paper.

### A.1 Proof of Theorem 19

To prove Theorem 19 (as well as later results), we will refer to *rank sets*.

**Definition 32.** Given a partial ranking of type $\gamma$, we denote the *rank set* occupied by $\Omega_i$ by $R_i^\gamma$. Note that $R_i^\gamma$ depends only on $\gamma$ and can be written as $R_1^\gamma = \{1, \ldots, \gamma_1\}$, $R_2^\gamma = \{\gamma_1 + 1, \ldots, \gamma_1 + \gamma_2\}, \ldots, R_r^\gamma = \{\sum_{i=1}^{r-1} \gamma_i + 1, \ldots, n\}$.

And we will refer to the following basic fact regarding rank sets:

**Proposition 33.** $\sigma \in S_\gamma \pi = \Omega_1 | \ldots | \Omega_r$ *if and only if for each $i$, $\sigma(\Omega_i) = R_i^\gamma$.*

*Proof.* (of Theorem 19) We use induction on the size of the itemset. The cases $n = 1, 2$ are trivial since every distribution on $S_1$ or $S_2$ factors riffle independently. We now consider the more general case of $n > 2$.

Fix a partial ranking $S_\gamma \pi = \Omega_1 | \Omega_2 | \ldots | \Omega_r$ of type $\gamma$ and a binary partition of the item set into subsets $A$ and $B$. We will show that the indicator function $\delta_{S_\gamma \pi}$ factors as:

$$\delta_{S_\gamma \pi}(\sigma) = m(\tau_{AB}(\sigma)) \cdot f(\phi_A(\sigma)) \cdot g(\phi_B(\sigma)), \tag{A.1}$$

where factors $m$, $f$ and $g$ are the indicator functions for the set of consistent interleavings, $[S_\gamma \sigma]_{AB}$, and the sets of consistent relative rankings, $[S_\gamma \sigma]_A$ and $[S_\gamma \sigma]_B$, respectively. If Equation A.1 is true, then we will have shown that $\delta_{S_\gamma \pi}$ must decompose with respect to the top layer of $H$. To show that $\delta_{S_\gamma \pi}$ decomposes hierarchically, we must also show that the relative ranking factors $f_A$ and $g_B$ decompose with respect to $H_A$ and $H_B$, the subhierarchies over the item sets $A$ and $B$. To establish this second step (assuming that Equation A.1 holds), note that $f_A$ and $g_B$ are indicator functions for the restricted partial rankings, $[S_\gamma \sigma]_A$ and $[S_\gamma \sigma]_B$, which themselves are partial rankings over smaller item sets $A$ and $B$. The inductive hypothesis (and the fact that $A$ and $B$ are assumed to be strictly smaller sets than $\Omega$) then shows that the functions $f_A$ and $g_B$ both factor according to their respective subhierarchies.

We now turn to establishing Equation A.1. It suffices to prove that the following two statements are equivalent:

I. The ranking $\sigma$ is consistent with the partial ranking $S_\gamma \pi$ (i.e., $\sigma \in S_\gamma \pi$).

II. The following three conditions hold:

    (a) The interleaving $\tau_{AB}(\sigma)$ is consistent with $S_\gamma \pi$   (i.e., $\tau_{AB}(\sigma) \in [S_\gamma \pi]_{AB}$), and

    (b) The relative ranking $\phi_A(\sigma)$ is consistent with $S_\gamma \pi$   (i.e., $\phi_A(\sigma) \in [S_\gamma \pi]_A$), and

    (c) The relative ranking $\phi_B(\sigma)$ is consistent with $S_\gamma \pi$   (i.e., $\phi_B(\sigma) \in [S_\gamma \pi]_B$).

- *(I ⇒ II):* We first show that $\sigma \in S_\gamma \pi$ implies conditions (a), (b) and (c).

  (a) If $\sigma \in S_\gamma \pi$, then for each $i$,

  $$|j \in R_i^\gamma : \tau_{AB}(j) = A| = |j \in R_i^\gamma : \sigma^{-1}(j) \in A|, \qquad \text{(by Definition 2)}$$
  $$= |k \in \Omega_i : k \in A|, \qquad \text{(by Proposition 33)}$$
  $$= |\Omega_i \cap A|.$$

  The same argument (replacing $A$ with $B$) shows that for each $i$, we have $|j \in R_i^\gamma : \tau_{AB}(j) = B| = |\Omega_i \cap B|$. These two conditions (by Definition 17) show that $\tau_{AB}$ is consistent with $S_\gamma \pi$.

  (b) If $\sigma \in S_\gamma \pi$, then (by Definition 14) $\sigma$ ranks items in $\Omega_i$ before items in $\Omega_j$ for any $i < j$. Intersecting each $\Omega_i$ with $A$, we also see that $\sigma$ ranks any item in $\Omega_i \cap A$ before any item in $\Omega_j \cap A$ for all $i, j$. By Definition 2, $\phi_A(\sigma)$ also ranks any item in $\Omega_i \cap A$ before any item in $\Omega_j \cap A$ for all $i, j$. And finally by Definition 16 again, we see that $\phi_A(\sigma)$ is consistent with the partial ranking $S_\gamma \pi$.

  (c) (Same argument as (b)).

- *(II ⇒ I):* We now assume conditions (a), (b), and (c) to hold, and show that $\sigma \in S_\gamma \pi$. By Proposition 33 it is sufficient to show that if an item $k \in \Omega_i$, then $\sigma(k) \in R_i^\gamma$. To prove this claim, we show by induction on $i$ that if an item $k \in \Omega_i \cap A$, then $\sigma(k) \in R_i^\gamma$ (and similarly if $k \in \Omega_i \cap B$, then $\sigma(k) \in R_i^\gamma$).

  *Base case.* In the base case ($i = 1$), we assume that $k \in \Omega_1 \cap A$, and the goal is to show that $\sigma(k) \in R_1$. By condition (a), we have that $\tau_{AB}(\sigma) \in [S_\gamma \pi]_{AB}$. By Definition 17, this means that: $|\Omega_1 \cap A| = \{j \in R_1 : [\tau_{AB}(\sigma)](j) = A\} = \{j \in R_1 : \sigma^{-1}(j) \in A\}$. In words, there are $m = |\Omega_1 \cap A|$ items from $A$ which lie in rank set $R_1 = \{1, \ldots, \gamma_1\}$. To show that an item $k \in A$ maps to a rank in $R_1$, we now must show that in the relative ranking of elements in $A$, $k$ is among the first $m$. By condition (b), $\phi_A(\sigma) \in [S_\gamma \pi]_A$, implying that the item subset $\Omega_1 \cap A$ occupies the first $m$ positions in the relative ranking of $A$. Since $k \in \Omega_1 \cap A$, item $k$ is among the first $m$ items ranked by $\phi_A(\sigma)$ and therefore $\sigma(k) \in R_1$. A similar argument shows that $k \in \Omega_1 \cap B$ implise that $\sigma(k) \in R_1$.

  *Inductive case.* We now show that if $k \in \Omega_i \cap A$, then $\sigma(k) \in R_i$. By condition (b), $\phi_A(\sigma) \in [S_\gamma \pi]_A$, implying that the item subset $\Omega_i \cap A$ (and hence, item $k$) occupies the first $m = |\Omega_i \cap A|$ positions in the relative ranking of $A$ beyond the items $\cup_{j=1}^{i-1}(\Omega_j \cap A)$. By the inductive hypothesis and mutual exclusivity, these items, together with $\cup_{j=1}^{i-1}(\Omega_j \cap B)$ occupy ranks $\cup_{j=1}^{i-1} R_j$, and therefore $\sigma(k) \in R_\ell$ for some $\ell \geq i$. On the other hand, condition (a) assures us that $|\Omega_i \cap A| = \{j \in R_i : \sigma^{-1}(j) \in A\}$ — or in other words, that the ranks in $R_i$ are occupied by exactly $m$ items of $A$. Therefore, $\sigma(k) \in R_i$. Again, a similar argument shows that $k \in \Omega_i \cap B$ implies that $\sigma(k) \in R_i$.

  □

## A.2 The PSPAN of a Set is Always a Partial Ranking

To reason about the PSPAN of a set of rankings, we first introduce some basic concepts regarding the combinatorics of partial rankings. The collection of partial rankings over $\Omega$ forms a *partially ordered set (poset)* where $S_{\gamma'}\pi' \prec S_\gamma\pi$ if $S_\gamma\pi$ can be obtained from $S_{\gamma'}\pi'$ by dropping vertical lines. For example, on $S_3$, we have that $1|2|3 \prec 12|3$. The *Hasse diagram* is the graph in which each node corresponds to a partial ranking and a node $x$ is connected to node $y$ via an edge if $x \prec y$ and there exists no partial ranking $z$ such that $x \prec z \prec y$ (see Lebanon & Mao, 2008). At the top of the Hasse diagram is the partial ranking $1, 2, \ldots, n$ (i.e., all of $S_\Omega$) and at the bottom of the Hasse diagram lie the full rankings. See Figure 13 for an example of the partial ranking lattice on $S_3$.

**Lemma 34.** *[Lebanon & Mao, 2008] Given any two partial rankings $S_\gamma\pi$, $S_{\gamma'}\pi'$, there exists a unique supremum of $S_\gamma\pi$ and $S_{\gamma'}\pi'$ (a node $S_{\gamma_{sup}}\pi_{sup}$ such that $S_\gamma\pi \prec S_{\gamma_{sup}}\pi_{sup}$ and $S_{\gamma'}\pi' \prec S_{\gamma_{sup}}\pi_{sup}$, and any other such node is greater than $S_{\gamma_{sup}}\pi_{sup}$). Similarly, there exists a unique infimum of $S_\gamma\pi$ and $S_{\gamma'}\pi'$.*

**Lemma 35.** *Given two partial rankings $S_\gamma\pi$, $S_{\gamma'}\pi'$, the relation $S_{\gamma'}\pi' \subset S_\gamma\pi$ holds if and only $S_\gamma\pi$ lies above $S_{\gamma'}\pi'$ in the Hasse diagram.*

*Proof.* If $S_\gamma\pi$ lies above $S_{\gamma'}\pi'$ in the Hasse diagram, then $S_{\gamma'}\pi' \subset S_\gamma\pi$ is trivial since $S_\gamma\pi$ can be obtained by dropping vertical bars of $S_{\gamma'}\pi'$. Now given that $S_\gamma\pi$ *does not* lie above $S_{\gamma'}\pi'$, we would like to show that $S_{\gamma'}\pi' \not\subset S_\gamma\pi$. Let $S_{\gamma_{inf}}\pi_{inf}$ be the unique infimum of $S_\gamma\pi$ and $S_{\gamma'}\pi'$ as guaranteed by Lemma 34. By the definition of the Hasse diagram, both $S_\gamma\pi$ and $S_\gamma\pi$ can be obtained by 'dropping' verticals from the vertical bar representation of $S_{\gamma_{inf}}\pi_{inf}$. Since $S_\gamma\pi$ does not lie above $S_{\gamma'}\pi'$, there must be a vertical bar that was dropped by $S_{\gamma'}\pi'$ which was not dropped by $S_\gamma\pi$ (if there does not exist such a bar, then $S_{\gamma'}\pi' \subset S_\gamma\pi$), and hence there must exist a pair of items $i, j$ separated by a single vertical bar in $S_\gamma\pi$ but unseparated in $S_{\gamma'}\pi'$. Therefore there exists $\sigma \in S_{\gamma'}\pi'$ such that $\sigma(j) < \sigma(i)$ even though there exists no such $\sigma \in S_\gamma\pi$. We conclude that $S_{\gamma'}\pi' \not\subset S_\gamma\pi$. $\qquad\square$

**Lemma 36** (Lemma 23 in main body). *For any $X \subset S_n$, $\mathrm{PSPAN}(X)$ is a partial ranking.*

*Proof.* Consider any subset $X \subset S_n$. A partial ranking containing every element in $X$ must be an upper bound of every element of $X$ in the Hasse diagram by Lemma 35. By Lemma 34, there must exist a unique least upper bound (supremum) of $X$, $S_{\gamma_{sup}}\pi_{sup}$, such that for any common upper bound $S_\gamma\pi$ of $X$, $S_\gamma\pi$ must also be an ancestor of $S_{\gamma_{sup}}\pi_{sup}$ and hence $S_{\gamma_{sup}}\pi_{sup} \subset S_\gamma\pi$. We therefore see that any partial ranking containing $X$ must be a superset of $S_{\gamma_{sup}}\pi_{sup}$. On the other hand, $S_{\gamma_{sup}}\pi_{sup}$ is itself a partial ranking containing $X$. Since $\mathrm{PSPAN}(X)$ is the intersection of partial rankings containing $X$, we have $\mathrm{PSPAN}(X) = S_{\gamma_{sup}}\pi_{sup}$ and therefore that $\mathrm{PSPAN}(X)$ must be a partial ranking. $\qquad\square$

### A.3 Proofs for the Claim that $\mathrm{RSPAN}(X) = \mathrm{PSPAN}(X)$

To simplify the notation in some of the remaining proofs, we introduce the following definition.

**Definition 37** (Ties). *Given a partial ranking $S_\gamma\pi = \Omega_1 | \ldots | \Omega_r$, we say that items $a_1$ and $a_2$ are tied (written $a_1 \sim a_2$) with respect to $S_\gamma\sigma$ if $a_1, a_2 \in \Omega_i$ for some $i$.*

The following basic properties of the tie relation are straightforward.

**Proposition 38.**

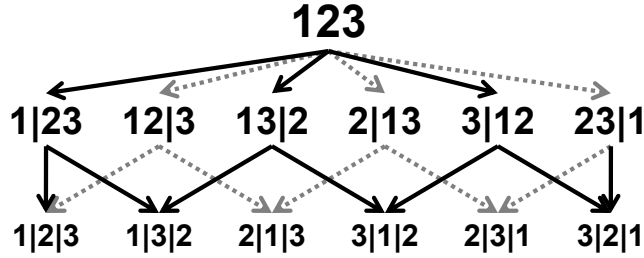

Figure 13: The Hasse diagram for the lattice of partial rankings on $S_3$.

I. *With respect to a fixed partial ranking $S_\gamma\pi$, the tie relation, $\sim$, is an equivalence relation on the item set (i.e., is reflexive, symmetric and transitive).*

II. *If there exist $\sigma, \sigma' \in S_\gamma\pi$ which disagree on the relative ranking of items $a_1$ and $a_2$, then $a_1 \sim a_2$ with respect to $S_\gamma\pi$.*

III. *If $S_\gamma\pi \prec S_{\gamma'}\pi'$, and $a_1 \sim a_2$ with respect to $S_\gamma\pi$, then $a_1 \sim a_2$ with respect to $S_{\gamma'}\pi'$.*

IV. *If $a_1 \sim a_2$ with respect to $S_\gamma\pi$, and $\sigma(a_1) < \sigma(a_3) < \sigma(a_2)$ for some item $a_3 \in \Omega$ and some $\sigma \in S_\gamma\pi$, then $a_1 \sim a_2 \sim a_3$.*

**Proposition 39.** *Given a set of rankings $X$ as input, Algorithm 2 outputs* pspan($X$).

*Proof.* We prove three things, which together prove the proposition: (1) that the algorithm terminates, (2) that at each stage the elements of $X$ are contained in pspan($X$), and (3) that upon termination, pspan($X$) is contained in each element of $X$.

1. First we note that the algorithm must terminate in finitely many iterations of the while loop since at each stage at least one vertical bar is removed from a partial ranking, and when all of the vertical bars have been removed from the elements of $X$, there are no disagreements on relative ordering.

2. We now show that at any stage in the algorithm, every element of $X_t$ is a subset of the pspan($X$). At initialization, of course, if $S_\gamma\pi \in X_0$, then it is simply a singleton set consisting of an element of $X$, and therefore $S_\gamma\pi \subset$ pspan($X$).

   Suppose now that $S_\gamma\pi \subset$ pspan($X$) for every $S_\gamma\pi \in X_t$. If $S_\gamma\pi$ is replaced by $S_{\tilde\gamma}\tilde\pi$ in $X_{t+1}$, then we want to show that $S_{\tilde\gamma}\tilde\pi \subset$ pspan($X$) as well. From Algorithm 2, for some $j$, if $S_\gamma\pi = \Omega_1|\dots|\Omega_j|\Omega_{j+1}|\dots|\Omega_r$, $S_{\tilde\gamma}\tilde\pi$ can be written as $\Omega_1|\dots|\Omega_j \cup \Omega_{j+1}|\dots|\Omega_r$, where the vertical bar between $\Omega_j$ and $\Omega_{j+1}$ is deleted due to the existence of some partial ranking in $X_t$, $S_{\gamma'}\pi' \in X_t$ which disagrees with $S_\gamma\pi$ on the relative ordering of items $a_1, a_2$ on opposite sides of the bar. Since $S_\gamma\pi$ and $S_{\gamma'}\pi'$ are both subsets of pspan($X$) by assumption, we know that $a_1 \sim a_2$ with respect to pspan($X$) (Proposition 38, II). Suppose now that $a_1 \in \Omega_i$ and $a_2 \in \Omega_{i'}$. Then for any $x \in \Omega_i$ and $y \in \Omega_{i'}$, we have $x \sim a_1$ and $y \sim a_2$ with respect to pspan($X$) by (III) of Proposition 38. Moreover, by (I, transitivity), we see that $x \sim y$ with respect to pspan($X$). for any two elements of $\Omega_i$ and $\Omega_{i'}$. By (IV) of Proposition 38, all the items lying in $\Omega_i, \Omega_{i+1}, \dots, \Omega_{i'}$ are thus tied with respect to pspan($X$) and therefore removing any bar between items $a_1$ and $a_2$ (producing, for example, $S_{\tilde\gamma}\tilde\pi$) results in a partial ranking which is a subset of pspan($X$).

3. Finally, upon termination, if some ranking $\sigma \in X$ is not contained in some element $S_\gamma \pi \in X_t$, then there would exist two items $a_1, a_2$ whose relative ranking $\sigma$ and $S_\gamma \pi$ disagree upon, which is a contradiction. Therefore, every element $S_\gamma \pi \in X_t$ contains every element of $X$ and thus $\text{PSPAN}(X) \subset S_\gamma \pi$ for every $S_\gamma \pi \in X_t$.

$\square$

**Lemma 40.** *Let $S_\gamma \pi = \Omega_1 | \ldots | \Omega_i | \Omega_{i+1} | \ldots | \Omega_k$ be a partial ranking on item set $\Omega$, and $S_{\gamma'} \pi' = \Omega_1 | \ldots | \Omega_i \cup \Omega_{i+1} | \ldots | \Omega_k$, the partial ranking in which the sets $\Omega_i$ and $\Omega_{i+1}$ are merged. Let $a_1 \in \cup_{j=1}^i \Omega_j$ and $a_2 \in \cup_{j=i+1}^k \Omega_j$. If $\mathcal{O}$ is any element of $\mathfrak{C}$ such that $S_\gamma \pi \subset \mathcal{O}$ and there additionally exists a ranking $\tilde{\pi} \in \mathcal{O}$ which disagrees with $S_\gamma \pi$ on the relative ordering of $a_1, a_2$, then $S_{\gamma'} \pi' \subset \mathcal{O}$.*

*Proof.* We will fix a completely decomposable $\mathcal{O}$ and again work with $h$, the indicator distribution corresponding to $\mathcal{O}$. Let $\sigma \in S_{\gamma'} \pi'$. To prove the lemma, we need to establish that $h(\sigma) > 0$. Let $\sigma^0$ be any element of $S_\gamma \pi$ such that $\sigma^0(k) = \sigma(k)$ for all $k \in \Omega \backslash (\Omega_i \cup \Omega_{i+1})$. Since $S_\gamma \pi \subset \text{supp}(h)$ by assumption, we have that $h(\sigma^0) > 0$.

Since $\sigma^0$ and $\sigma$ match on all items except for those in $\Omega_i \cup \Omega_{i+1}$, there exists a sequence of rankings $\sigma^0, \sigma^1, \sigma^2, \ldots, \sigma^m = \sigma$ such that adjacent rankings in this sequence differ only by a pairwise exchange of items $b_1, b_2 \in \Omega_i \cup \Omega_{i+1}$. We will now show that at each step along this sequence, $h(\sigma^t) > 0$ implies that $h(\sigma^{t+1}) > 0$, which will prove that $h(\sigma) > 0$. Suppose now that $h(\sigma^t) > 0$ and that $\sigma^t$ and $\sigma^{t+1}$ differ only by the relative ranking of items $b_1, b_2 \in \Omega_i \cup \Omega_{i+1}$ (without loss of generality, we will assume that $\sigma^t(b_2) < \sigma^t(b_1)$ and $\sigma^{t+1}(b_1) < \sigma^{t+1}(b_2)$).

The idea of the following paragraph is to use the previous lemma (Lemma 28) to prove that $\sigma^{t+1}$ has positive probability and to do so, it will be necessary to argue that there exists some ranking $\sigma'$ such that $h(\sigma') > 0$ and $\sigma'(b_1) < \sigma'(b_2)$ (i.e., $\sigma'$ disagrees with $\sigma^t$ on the relative ranking of $b_1, b_2$). Let $\omega$ be any element of $S_\gamma \pi$. If $a_1 \in \Omega_i$, rearrange $\omega$ such that $a_1$ is ranked first among elements of $\Omega_i$. If $a_2 \in \Omega_{i+1}$, further rearrange $\omega$ such that $a_2$ is ranked last among elements of $\Omega_{i+1}$. Note that $\omega$ is still an element of $S_\gamma \pi$ after the possible rearrangements and therefore $h(\omega) > 0$. We can assume that $\omega(b_2) < \omega(b_1)$ since otherwise we will have shown what we wanted to show. Thus the relative ordering of $a_1, a_2, b_1, b_2$ within $\omega$ is $a_1 | b_2 | b_1 | a_2$. Note that we treat the case where the items $a_1, a_2, b_1, b_2$ are distinct, but the same argument follows in the cases when $a_1 = b_2$ or $a_2 = b_1$.

Now since $\tilde{\pi}$ disagrees with $S_\gamma \pi$ on the relative ordering of $a_1, a_2$ by assumption (and hence disagrees with $\omega$), we apply Lemma 28 to conclude that swapping the relative ordering of $a_1, a_2$ within $\omega$ (obtaining $a_2 | b_2 | b_1 | a_1$) results in a ranking, $\omega'$, such that $h(\omega') > 0$. Finally, observe that $\omega$ and $\omega'$ must now disagree on the relative ranking of $a_2, b_2$, and invoking Lemma 28 again shows that we can swap the relative ordering of $a_2, b_2$ within $\omega$ (obtaining $a_1 | a_2 | b_1 | b_2$) to result in a ranking $\sigma'$ such that $h(\sigma') > 0$. This element $\sigma'$ ranks $b_1$ before $b_2$, which is what we wanted to show.

We have shown that there exist rankings which disagree on the relative ordering of $b_1$ and $b_2$ with positive probability under $h$. Again applying Lemma 28 shows that we can swap the relative ordering of items $b_1, b_2$ within $\sigma^t$ to obtain $\sigma^{t+1}$ such that $h(\sigma^{t+1}) > 0$, which concludes the proof. $\square$

### A.4 Uniformity of $\tilde{\mathfrak{C}}$ Functions Over a Partial Ranking

We have thus far shown that any element of $\tilde{\mathfrak{C}}$ must be supported on some partial ranking. In the following, we show that (up to a certain class of exceptions), such an element must assign uniform probability to all members of this partial ranking.

**Theorem 41.** *If $h$ is any completely decomposable function supported on a partial ranking $S_\gamma \pi = \Omega_1 | \ldots | \Omega_r$ where $|\Omega_i| \neq 2$ for all $i = 1, \ldots, r$, then $h$ is uniform on $S_\gamma \pi$ (i.e., $h(\sigma) = \frac{1}{\prod_i |\Omega_i|}$ for all $\sigma \in S_\gamma \pi$).*

To establish Theorem 41, we must establish two supporting results: (1) Lemma 42 which factors $h$ into $r$ smaller completely decomposable functions, each of which is nonzero everywhere on its domain, and (2) Theorem 43 which establishes uniformity for any completely decomposable function which is nonzero everywhere on its domain.

**Lemma 42.** *Any completely decomposable function, $h$, supported on the partial ranking $S_\gamma \pi = \Omega_1 | \ldots | \Omega_r$, must factor as: $h(\sigma) = \prod_{i=1}^{r} h(\sigma(\Omega_i))$, where each factor distribution $h(\sigma(\Omega_i))$ is itself a completely decomposable function on $S_{\Omega_i}$.*

*Proof.* Since $h$ is completely decomposable, we have that $\sigma(\Omega_i)$ is riffle independent of $\sigma(\Omega \backslash \Omega_i)$ for each $i$. Since $h$ is supported on the partial ranking $S_\gamma \pi = \Omega)_1 | \ldots | \Omega_r$, however, the interleaving of $\Omega_i$ with its complement is deterministic and therefore we conclude in fact that $\sigma(\Omega_i)$ is fully independent of $\sigma(\Omega \backslash \Omega_i)$. Since $\sigma(\Omega_i) \perp \sigma(\Omega \backslash \Omega_i)$ for each $i$, we have the factorization: $h(\sigma) = \prod_{i=1}^{r} h(\sigma(\Omega_i))$.

We now turn to establishing that each factor $h(\sigma(\Omega_i))$ is itself a completely decomposable observation. Fix $i = 1$ (without loss of generality) and consider any partition of the set $\Omega_1$ into subsets $A \cup B$. We would like to see that the sets $A$ and $B$ are riffle independent of each other with respect to $h(\sigma(\Omega_1))$. Since $h$ is assumed to be completely decomposable, we know that $A$ is riffle independent of its complement, $B \cup (\Omega \backslash \Omega_1)$. In other words, if $\tilde{B} = B \cup (\Omega \backslash \Omega_1)$, then the variables $\phi_A(\sigma), \tau_{A\tilde{B}}, \phi_{\tilde{B}}$ (the relative ranking of $A$, the interleaving of $A$ with all remaining items, and the relative ranking of all remaining items, respectively) are mutually independent. We then observe that (1) the interleaving of $A$ and $B$, $\tau_{AB}$, is a deterministic function of the interleaving of $\tau_{A\tilde{B}}$ and (2) the relative ranking of $B$, $\phi_B$, is a deterministic function of $\phi_{\tilde{B}}$, thus proving that $\phi_A, \tau_{AB}$ and $\phi_B$ are mutually independent and hence that $A$ and $B$ are riffle independent. $\square$

**Theorem 43.** *Let $h$ a completely decomposable function such that $h(\sigma) > 0$ for all $\sigma \in S_n$ for $n > 2$. Then for any two rankings $\sigma_1, \sigma_2$ which differ by a single transposition, we have $h(\sigma_1) = h(\sigma_2)$.*

Our proof strategy for Theorem 43 will involve examining the ratio between the two probabilities $h(\sigma_1)$ and $h(\sigma_2)$. We then define an operation transforming $\sigma_1$ and $\sigma_2$ into new rankings $\sigma_1'$ and $\sigma_2'$ such that the ratio between the rankings is preserved (i.e., $h(\sigma_1)/h(\sigma_2) = h(\sigma_1')/h(\sigma_2')$). By performing a sequence of such ratio-preserving operations, we show that:

$$\frac{h(\sigma_1)}{h(\sigma_2)} = \frac{h(\sigma_2)}{h(\sigma_1)},$$

from which Theorem 43 easily follows.

We will use two types of operations which transform a ranking into a new ranking: (1) changing the interleaving of two sets $A$ and $B$ within a ranking $\sigma$, and (2), changing the relative ranking of a set $A$ within a ranking $\sigma$. More precisely, given a ranking $\sigma$ and a partitioning of the item set into subsets $A$ and $B$, we can uniquely index $\sigma$ as a triplet $(\tau, \pi_A, \pi_B)$, where $\tau = \tau_{A,B}(\sigma)$, $\pi_A = \phi_A(\sigma)$, and $\pi_B = \phi_B(\sigma)$. The two operations are defined as follows:

1.  *Changing the interleaving of $A, B$ within $\sigma$ to $\tau'$:* yields the new ranking $\sigma'$ which is indexed by $(\tau', \pi_A, \pi_B)$.

2.  *Changing the relative ranking of $A$ (or $B$) within $\sigma$ to $\pi'_A$ (or $\pi'_B$):* yields the new ranking $\sigma'$ which is indexed by $(\tau, \pi'_A, \pi_B)$ [or $(\tau, \pi_A, \pi'_B)$].

If we use the above operations to obtain from $\sigma'_1$ and $\sigma'_2$, we are interested in conditions under which this transformation is ratio-preserving (i.e., $h(\sigma_1)/h(\sigma_2) = h(\sigma'_1)/h(\sigma'_2)$). The following lemma provides sufficient conditions for ratio-preservation.

**Lemma 44.** *Let $h$ be any completely decomposable function and consider $\sigma_1, \sigma_2 \in S_n$ such that $h(\sigma_2) > 0$. Then for any partitioning of the item set into subsets $A$ and $B$, we have:*

1.  *If $\sigma_1$ and $\sigma_2$ match on the interleaving of $A$ and $B$ (i.e., $\tau_{A,B}(\sigma_1) = \tau_{AB}(\sigma_2)$), then $\frac{h(\sigma_1)}{h(\sigma_2)} = \frac{h(\sigma'_1)}{h(\sigma'_2)}$, where $\sigma'_1$ and $\sigma'_2$ are formed by changing the interleaving of the sets $A$ and $B$ within $\sigma_1$ and $\sigma_2$ to be any new interleaving $\tau'$.*

2.  *If $\sigma_1$ and $\sigma_2$ match on the relative ranking of $A$ (or $B$) (i.e., $\phi_A(\sigma_1) = \phi_A(\sigma_2)$ (or $\phi_B(\sigma_1) = \phi_B(\sigma_2)$)), then $\frac{h(\sigma_1)}{h(\sigma_2)} = \frac{h(\sigma'_1)}{h(\sigma'_2)}$, where $\sigma'_1$ and $\sigma'_2$ are formed by changing the relative ranking of set $A$ (or $B$) within $\sigma_1$ and $\sigma_2$ to be any new relative ranking $\pi'_A$ (or $\pi'_B$).*

*Proof.* Since the proofs of parts 1 and 2 are nearly identical, we just prove part 1 here. Since $h \in \mathfrak{C}$, the sets $A$ and $B$ are riffle independent by assumption, and hence we have the factorizations:

$$\frac{h(\sigma_1)}{h(\sigma_2)} = \frac{m(\tau_1) \cdot f(\pi_1^A) \cdot g(\pi_2^B)}{m(\tau_2) \cdot f(\pi_2^A) \cdot g(\pi_2^B)}.$$

If $\sigma_1$ and $\sigma_2$ match on the interleaving of the sets $A$ and $B$, then we have that $\tau = \tau_1 = \tau_2$, and thus the interleaving terms, $m(\tau_1)$ and $m(\tau_2)$ are the same in both the numerator and denominator.

On the other hand, if we examine the ratio between $h(\sigma'_1)$ and $h(\sigma'_2)$, we also see that the interleaving terms must cancel:

$$\frac{h(\sigma_1)}{h(\sigma_2)} = \frac{m(\tau'_1) \cdot f(\pi_1^A) \cdot g(\pi_2^B)}{m(\tau'_2) \cdot f(\pi_2^A) \cdot g(\pi_2^B)}.$$

We therefore have that:

$$\frac{h(\sigma_1)}{h(\sigma_2)} = \frac{f(\pi_1^A) \cdot g(pi_1^B)}{f(\pi_2^A) \cdot g(\pi_2^B)} = \frac{h(\sigma'_1)}{h(\sigma'_2)}.$$

$\square$

Having now established Lemma 44, we turn to establishing three short claims (using the lemma) that will allow us to prove finally prove Theorem 43. It is interesting to note that we require $n > 2$ (strictly) in claim III below in which we swap the order of $i$ and $j$ in numerator and denominator. The third item $k$ in our proof below can be thought of as playing the role of a dummy variable analogous to the temporary storage variables that one might use in implementing a swap function. The necessity of this third item is precisely why our result does not hold in the special case that $n = 2$.

**Proposition 45.** *Let $h : S_n \to \mathbb{R}$ be a completely decomposable function with $n > 2$ with $h(\sigma) > 0$ for all $\sigma \in S_n$. We have the following equivalences (where in each of the below ratios, entries which have* not *been explicitly written out are assumed to match identically in both the numerator and denominator).*

*I.*
$$\frac{h(i|j|\dots|k|\dots)}{h(j|i|\dots|k|\dots)} = \frac{h(i|j|k|\dots)}{h(j|i|k|\dots)}.$$

*II.*
$$\frac{h(\dots|i|\dots|j|\dots)}{h(\dots|j|\dots|i|\dots)} = \frac{h(i|j|\dots)}{h(j|i|\dots)}.$$

*III.*
$$\frac{h(i|j|k|\dots)}{h(j|i|k|\dots)} = \frac{h(j|i|k|\dots)}{h(i|j|k|\dots)}.$$

*Proof.*

I. Equality holds in *I* since $\sigma_1$ and $\sigma_2$ match on the interleaving of the sets $A = \{k\}$ with $B = \Omega \backslash \{k\}$. Thus we can change the interleaving of $A$ and $B$ in both $\sigma_1$ and $\sigma_2$ so that item $k$ is inserted in rank 3 while preserving the ratio.

II. Equality holds in *II* since $\sigma_1$ and $\sigma_2$ match on the interleaving of the sets $A = \{i, j\}$ with $B = \Omega \backslash \{i, j\}$. Thus we can change the interleaving of $A$ and $B$ in both $\sigma_1$ and $\sigma_2$ so that items $i$ and $j$ occupy the first two ranks while preserving the ratio between $h(\sigma_1)$ and $h(\sigma_2)$.

III. In the following we use $\sigma_1$ and $\sigma_2$ to refer to the arguments in the numerator and denominator, respectively, of the preceding line.

$$\frac{h(i|j|k|\dots)}{h(j|i|k|\dots)} = \frac{h(i|k|j|\dots)}{h(k|i|j|\dots)}, \qquad \text{(since } \sigma_1, \sigma_2 \text{ match on the relative ranking of } \{j, k\})$$

$$= \frac{h(j|i|k|\dots)}{h(j|k|i|\dots)}, \qquad \text{(since } \sigma_1, \sigma_2 \text{ match on the interleaving of } \{j\} \text{ with } \Omega \backslash \{j\})$$

$$= \frac{h(i|j|k|\dots)}{h(i|k|j|\dots)}, \qquad \text{(since } \sigma_1, \sigma_2 \text{ match on the relative ranking of } \{i, j\})$$

$$= \frac{h(k|j|i|\dots)}{h(k|i|j|\dots)}, \qquad \text{(since } \sigma_1, \sigma_2 \text{ match on the relative ranking of } \{i, k\})$$

$$= \frac{h(j|i|k|\dots)}{h(i|j|k|\dots)}, \qquad \text{(since } \sigma_1, \sigma_2 \text{ match on the interleaving of } \{k\} \text{ with } \Omega \backslash \{k\}).$$

□

*Proof.* (of Theorem 43) We want to show that if two rankings differ by a single transposition, then they are assigned equal probability under $h$. Suppose then that $\sigma_2$ is obtained from $\sigma_1$ by swapping the ranks of items $i$ and $j$. Additionally, let $k$ be any item besides $i$ and $j$ (such an item must exist since $n > 2$). In the following, we use Proposition 45 to show that $h(\sigma_1)/h(\sigma_2) = h(\sigma_2)/h(\sigma_1)$. As before, entries which have *not* been explicitly written out are assumed to match identically in both the numerator and denominator.

$$\frac{h(\sigma_1)}{h(\sigma_2)} = \frac{h(\ldots|i|\ldots|j|\ldots)}{h(\ldots|j|\ldots|i|\ldots)} = \frac{h(i|j|\ldots)}{h(j|i|\ldots)}, \text{ (by Prop. 45, Part II)}$$

$$= \frac{h(i|j|\ldots|k|\ldots)}{h(j|i|\ldots|k|\ldots)} = \frac{h(i|j|k|\ldots)}{h(j|i|k|\ldots)}, \text{ (by Prop. 45, Part I)}$$

$$= \frac{h(j|i|k|\ldots)}{h(i|j|k|\ldots)}, \text{ (by Prop. 45, Part III)}$$

$$= \frac{h(j|i|\ldots|k|\ldots)}{h(i|j|\ldots|k|\ldots)}, \text{ (by Prop. 45, Part I)}$$

$$= \frac{h(j|i|\ldots)}{h(i|j|\ldots)} = \frac{h(\ldots|j|\ldots|i|\ldots)}{h(\ldots|i|\ldots|j|\ldots)}, \text{ (by Prop. 45, Part II)}$$

$$= \frac{h(\sigma_2)}{h(\sigma_1)}.$$

Since we have assumed $h(\sigma_1)$ and $h(\sigma_2) > 0$, we must conclude that $h(\sigma_1) = h(\sigma_2)$. □

Finally, we assemble all of our supporting results to prove Theorem 41.

*Proof. (of Theorem 41)* By Lemma 42, a completely decomposable function $h$ must factor as:

$$h(\sigma) = \prod_{i=1}^{r} h(\sigma(\Omega_i)), \tag{A.2}$$

where each factor distribution $h(\sigma(\Omega_i))$ is itself a completely decomposable function on $S_{\Omega_i}$. By assumption, $|\Omega_i| \neq 2$. If $|\Omega_i| = 1$, then its corresponding factor $h(\sigma(\Omega_i))$ must trivially be uniform. Otherwise, we have that $|\Omega_i| > 2$. In this latter case, we apply Theorem 43 to $h(\sigma(\Omega_i))$ to show that it must assign equal probability to any two rankings that differ by a single transposition. However, given any rankings $\sigma_1, \sigma_2 \in S_{\Omega_i}$, we can obtain a sequence of transpositions that transforms $\sigma_1$ into $\sigma_2$, and therefore, Theorem 43 in fact implies that the factor $h(\sigma(\Omega_i))$ is constant on all inputs. Having proved that each factor in Equation A.2 is constant, we conclude that $h$ must be constant on its support. □